\documentclass[a4paper,fleqn]{cas-dc}

\usepackage[authoryear]{natbib}

\usepackage{amssymb}
\usepackage{amsmath}
\usepackage[switch]{lineno}
\usepackage{graphicx}
\usepackage{textcomp}
\usepackage{xcolor}
\usepackage{hyperref}
\usepackage{natbib}
\usepackage{booktabs}
\usepackage{placeins}

\definecolor{increasecolor}{HTML}{1B7F3A}
\definecolor{decreasecolor}{HTML}{C62828}
\definecolor{nochangecolor}{HTML}{666666}
\newcommand{\gain}[1]{\textcolor{increasecolor}{\scriptsize$\uparrow\,#1$}}

\newcommand{\nochange}[1]{\textcolor{nochangecolor}{\scriptsize$\rightarrow\,#1$}}

\def\tsc#1{\csdef{#1}{\textsc{\lowercase{#1}}\xspace}}
\tsc{WGM}
\tsc{QE}
\tsc{EP}
\tsc{PMS}
\tsc{BEC}
\tsc{DE}

\begin{document}
\let\WriteBookmarks\relax
\def\floatpagepagefraction{1}
\def\textpagefraction{.001}

\shorttitle{CowCare: A Unified Visual Framework for Viewpoint-Robust Cattle Tracking and Spatiotemporal Heat-Stress Analysis}

\shortauthors{Liu et~al.}


\title [mode = title]{CowCare: A Unified Visual Framework for Viewpoint-Robust Cattle Tracking and Spatiotemporal Heat-Stress Analysis}

%
\author[1,2,3,4]{Xin Liu}[orcid=0009-0004-4307-9932]
\credit{Writing - original draft, Writing - review \& editing, Methodology, Investigation, Formal analysis, Data curation, Visualization, Validation, Resources}

\author[1,2,3,4]{Weiran Li}[]
\credit{Project administration, Writing - review \& editing} 

\author[1,2,3,4]{Minghao Zhu}[]
\credit{Validation, Data curation} 

\author[1,2,3,4]{Mina Han}[]
\credit{Formal analysis, Investigation} 

\author[1,2,3,4]{Zhenbo Li}[orcid=0000-0003-2914-1192]
\credit{Supervision, Resources, Investigation, Funding acquisition, Writing - review \& editing} 
\cormark[1]

\affiliation[1]{organization={College of Information and Electrical Engineering, China Agricultural University}, city={Beijing}, postcode={100083}, country={China}}
\affiliation[2]{organization={National Innovation Center for Digital Fishery, Ministry of Agriculture and Rural Affairs}, city={Beijing}, postcode={100083}, country={China}}
\affiliation[3]{organization={Key Laboratory of Smart Farming Technologies for Aquatic Animal and Livestock, Ministry of Agriculture and Rural Affairs}, city={Beijing}, postcode={100083}, country={China}}
\affiliation[4]{organization={Key Laboratory of Agricultural Informatization Standardization, Ministry of Agriculture and Rural Affairs}, city={Beijing}, postcode={100083}, country={China}}

\cortext[cor1]{Corresponding author. Email: lizb@cau.edu.cn}


\begin{abstract}
Reliable individual monitoring is essential for assessing dairy cattle welfare, particularly for early heat stress detection in Precision Livestock Farming (PLF). However, translating raw multi-camera videos into automated health diagnostics remains challenging due to disjointed tracking and recognition pipelines, as well as complex physical environments. Specifically, dense mutual occlusions and cross-camera viewpoint variations cause frequent identity fragmentation. Meanwhile, extracting subtle heat stress indicators, such as rapid breathing and flank undulations, is easily compromised by environmental noise (e.g., camera vibrations) and data scarcity in agricultural datasets. To bridge the gap between low-level visual features and high-level behavioral diagnostics, we propose CowCare, a unified framework that integrates viewpoint-robust tracking and spatiotemporal heat stress recognition. For tracking, our module, CowTrack, maintains cross-camera identity consistency using a View-Decoupled Aggregation Network (VDA-Net). Instead of relying on a static global representation, VDA-Net aggregates localized textures to handle heavy occlusions and employs a Viewpoint Decoupling Transformer (VDT) to filter out transient viewpoint variations. CowTrack further incorporates a causal online self-supervision mechanism for continuous adaptation to dynamic scenes. Subsequently, the extracted tracklets are processed by TSDF-Net, a fine-grained video understanding network. TSDF-Net utilizes a Temporal Stabilization and Aggregation Module (TSAM) to suppress high-frequency jitter, alongside a Diffusion Semantic Prior (DSP) that leverages frozen pre-trained diffusion semantics to amplify weak behavioral signals while mitigating overfitting. Extensive evaluations on two real-world farm benchmarks validate the effectiveness of our framework. On the MVCow dataset, CowTrack achieves 83.48\% HOTA and 87.72\% IDF1 for global multi-object tracking. On the HSCow dataset, TSDF-Net attains 87.72\% Top-1 accuracy, demonstrating that CowCare provides a robust end-to-end solution for automated livestock health monitoring.
\end{abstract}




\maketitle

\section{Introduction}

Continuous monitoring of individual animal health and behavior is a cornerstone of Precision Livestock Farming (PLF), which aims to enhance both productivity and animal welfare~\citep{shu2023predicting}. Among various health threats, heat stress presents a critical challenge in commercial dairy farming, leading to physiological disruptions that impair milk production, reproductive performance, and immune function~\citep{chapman2023deep, chen2024effects}. Traditionally, assessing heat stress has relied on environmental metrics, such as the Temperature-Humidity Index (THI) and microclimatic conditions~\citep{chapman2023deep, leliveld2024real, zanchi2025influence}, or labor-intensive manual observations like panting scores~\citep{russello2024video}. Contact-based physiological measurements, such as wearable sensors or rectal temperature probes, have also been utilized~\citep{rutten2013sensors}. However, these methods are often costly, difficult to deploy at scale, and may inadvertently induce additional stress in the animals~\citep{shu2023predicting}. Consequently, computer vision-based monitoring and deep learning methodologies have emerged as a promising, non-invasive alternative that enables continuous, large-scale assessment of cattle well-being~\citep{wang2023computer}.

To convert raw barn videos into reliable individual-level health diagnostics, a visual monitoring system must couple Multi-Object Tracking (MOT) for long-term identity preservation with spatiotemporal action recognition for behavior analysis~\citep{zheng2023prunedyolo, zheng2023yolo}. Recently, MOT has been widely adopted in precision livestock farming for diverse species, including pigs~\citep{lu2024orp, chen2025pkl} and poultry~\citep{li2022bgd, JIANG2025110640}. While tracking-by-detection paradigms driven by kinematic predictions~\citep{zhang2022bytetrack, stanojevic2024boosttrack++} and sophisticated spatiotemporal architectures like 3D CNNs~\citep{feichtenhofer2019slowfast} and Transformers~\citep{vaswani2017attention,dulal2025mhaff} have significantly advanced general computer vision, deploying such an end-to-end algorithmic pipeline in real-world farm environments remains challenging. Existing studies typically treat tracking and behavior recognition as disjoint tasks, which complicates the transition from pixel-level features to semantic diagnostics in uncalibrated, non-overlapping multi-camera networks.

\begin{figure*}
	\centering
	\includegraphics[width=0.95\linewidth]{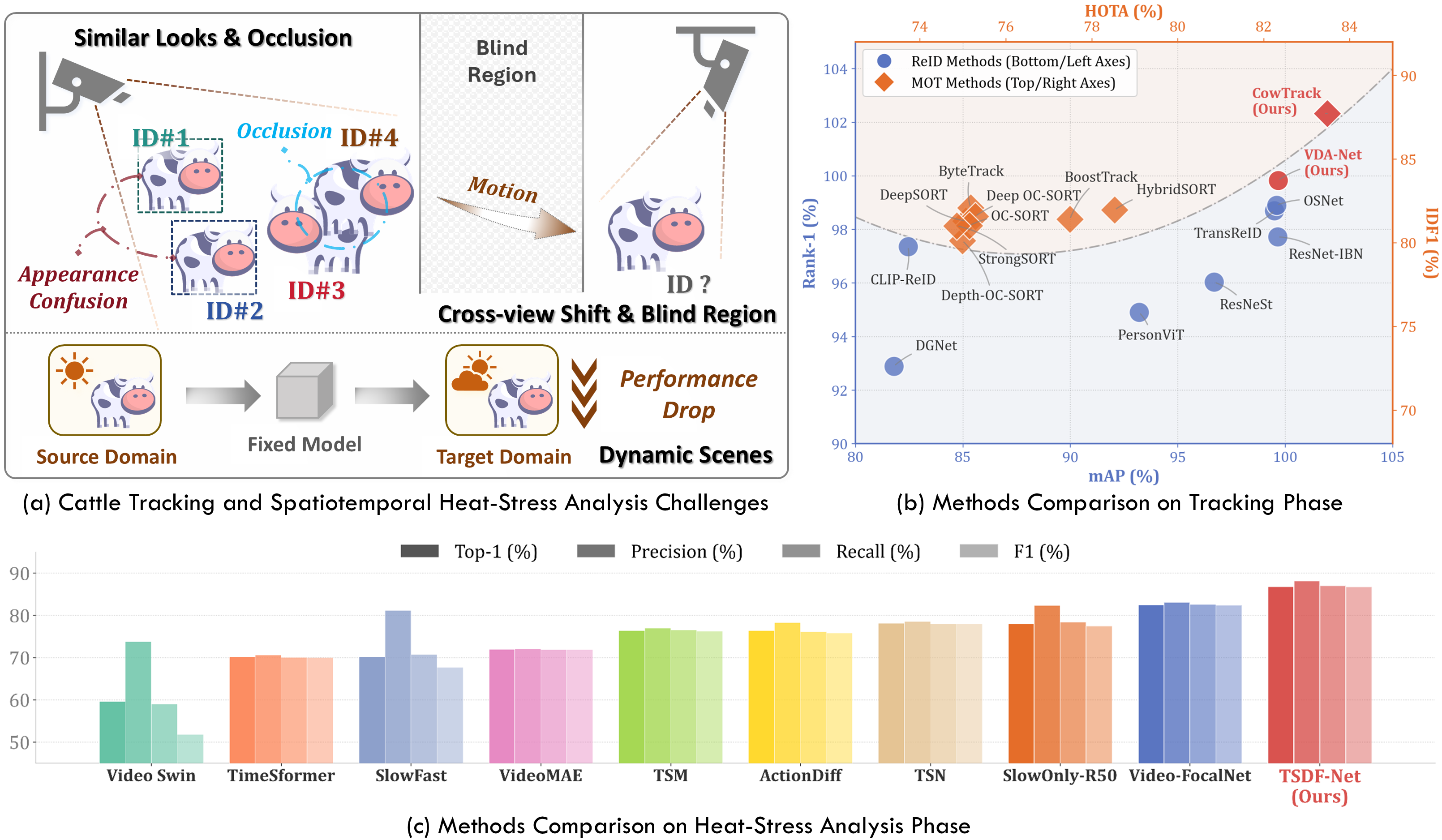}
	\caption{Overview of the visual challenges and the comprehensive performance of our proposed methods. (a) Critical challenges in cattle tracking and heat-stress analysis, including appearance confusion under dense occlusion, identity fragmentation across non-overlapping views, and domain gaps in dynamic scenes. (b) Performance comparison of our CowTrack and VDA-Net against state-of-the-art MOT and ReID methods. (c) Performance comparison of our TSDF-Net against existing spatiotemporal video understanding models for fine-grained heat-stress recognition.}
	\label{fig:cover}
\end{figure*}

Beyond this system-level disjointness, practical deployment is severely hindered by critical visual challenges, as illustrated in Fig.~\ref{fig:cover}(a). First, preserving identities in dense herds across multi-camera networks is difficult. In commercial barns, traditional geometric tracking cues inevitably fail due to prolonged mutual occlusion and nonlinear animal movements~\citep{cao2023observation}, necessitating robust appearance-based re-identification (ReID). Inspired by advancements in general person and object re-identification~\citep{wu2016personnet, ye2024transformer}, recent livestock ReID methods have progressed from shallow networks for global coat patterns~\citep{li2017automatic, andrew2017visual, bhole2019computer} to deep representations, multi-view fusions, and domain adaptation~\citep{chen2022holstein, gao2022label, dubourvieux2023cumda, sharma2025universal}. Specifically for monitoring, various models have been developed to address long-term tracking challenges across cattle and pigs~\citep{odo2025re, wang2023shufflenet, yu2025multicamcows, gou2026foonet}. However, static representation models still struggle to isolate an animal's intrinsic identity from severe viewpoint distortions across non-overlapping cameras~\citep{liu2025hanwooreid, zhang2024view}. This vulnerability to dynamic lighting and shifting herd compositions frequently leads to identity switches and fragmented trajectories. Second, extracting fine-grained behavioral semantics under environmental noise is a complex task. While existing PLF methods effectively classify macroscopic activities (e.g., standing, walking) using temporal sequence networks or customized 3D CNNs, the visual manifestations of early heat stress, such as rapid breathing and subtle flank undulations, are minute and localized. These subtle signals are easily obscured by high-frequency physical noise, notably camera vibrations induced by barn cooling fans and unpredictable illumination shifts. Furthermore, the scarcity of expert-annotated veterinary video data for specialized livestock behaviors restricts the feature representation capabilities of standard deep networks, increasing the risk of overfitting.

To address these intertwined challenges, we propose CowCare, a unified visual framework for viewpoint-robust cattle tracking and spatiotemporal heat stress recognition. By seamlessly linking stable identity association with fine-grained behavior analysis, CowCare establishes a comprehensive pipeline that directly translates multi-camera video streams into automated heat stress diagnostics. For identity preservation, we introduce CowTrack, an MOT framework driven by a novel appearance model named View-Decoupled Aggregation Network (VDA-Net). VDA-Net utilizes a localized aggregation strategy to capture visible textures under severe occlusion and employs a Viewpoint Decoupling Transformer (VDT) to separate intrinsic identity embeddings from transient, viewpoint-dependent variations. To break the bottleneck of static deployment, we further introduce a causal online self-supervised adaptation strategy, enabling the tracker to update its representations dynamically without manual annotations. For fine-grained recognition amidst environmental noise, the extracted tracklets are processed by a fine-grained video understanding network, TSDF-Net. TSDF-Net mitigates physical noise via a Temporal Stabilization and Aggregation Module (TSAM), which smooths inter-frame jitter and adaptively down-weights corrupted frames. To compensate for data scarcity, we introduce a Diffusion Semantic Prior (DSP). Inspired by the rich, generalized motion priors encapsulated in generative diffusion models, DSP leverages a frozen pre-trained video diffusion encoder to integrate generalized motion semantics into the network, amplifying subtle dynamic traits of heat stress.

The main contributions of this work are summarized as follows:
\begin{itemize}
    \item We propose an end-to-end framework that integrates viewpoint-robust tracking and anti-jitter spatiotemporal analysis, facilitating continuous, non-invasive health monitoring of individual cattle in uncalibrated multi-camera networks.
    \item We design the VDA-Net appearance model to tackle severe occlusion and cross-camera variations via localized texture aggregation and viewpoint decoupling. Coupled with an online self-supervised adaptation mechanism, CowTrack achieves robust identity preservation in dynamic environments.
    \item We introduce a fine-grained spatiotemporal network that suppresses high-frequency physical vibrations via TSAM. Additionally, we integrate frozen Diffusion Semantic Priors (DSP) to overcome representation limitations inherent in small-scale agricultural datasets.
    \item We construct two real-world commercial farm benchmarks (MVCow and HSCow) and conduct comprehensive evaluations. As compared in Fig.~\ref{fig:cover}(b) and (c), the proposed CowCare framework demonstrates state-of-the-art performance. Notably, CowTrack achieves 83.48\% HOTA for robust tracking, while TSDF-Net attains 87.72\% Top-1 accuracy for heat-stress analysis, establishing strong baselines for automated livestock monitoring.
\end{itemize}

\section{Materials and Methods}
\label{sec:framework}

\begin{figure*}
	\centering
	\includegraphics[width=0.80\linewidth]{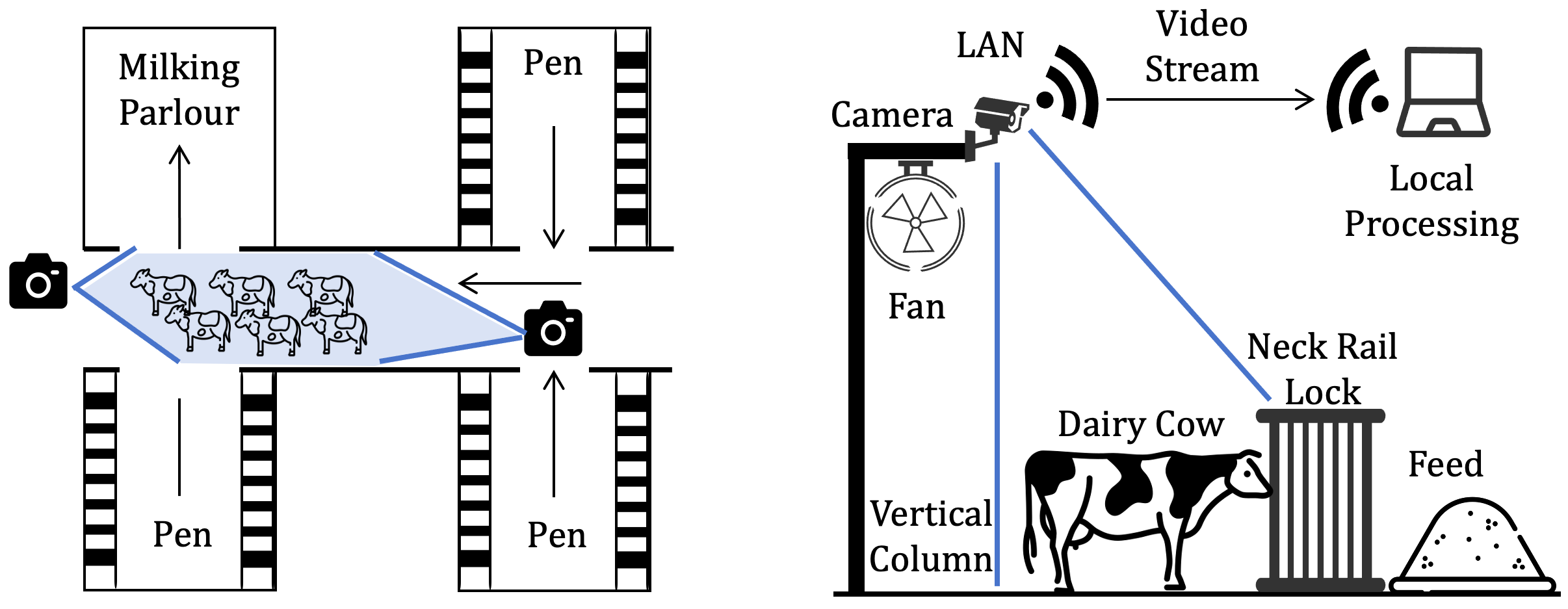}
	\caption{Equipment deployment and the actual farming environment for dataset collection.}
	\label{fig:hscow_capture}
\end{figure*}

\subsection{Dataset Construction}
To validate the proposed framework, we constructed two complementary benchmarks from commercial dairy farms, as detailed in Table~\ref{tab:data_collection}: MVCow for cross-camera tracking, and HSCow for heat stress recognition (see Fig.~\ref{fig:hscow_capture} for equipment deployment).

\begin{figure}
	\centering
	\includegraphics[width=0.95\linewidth]{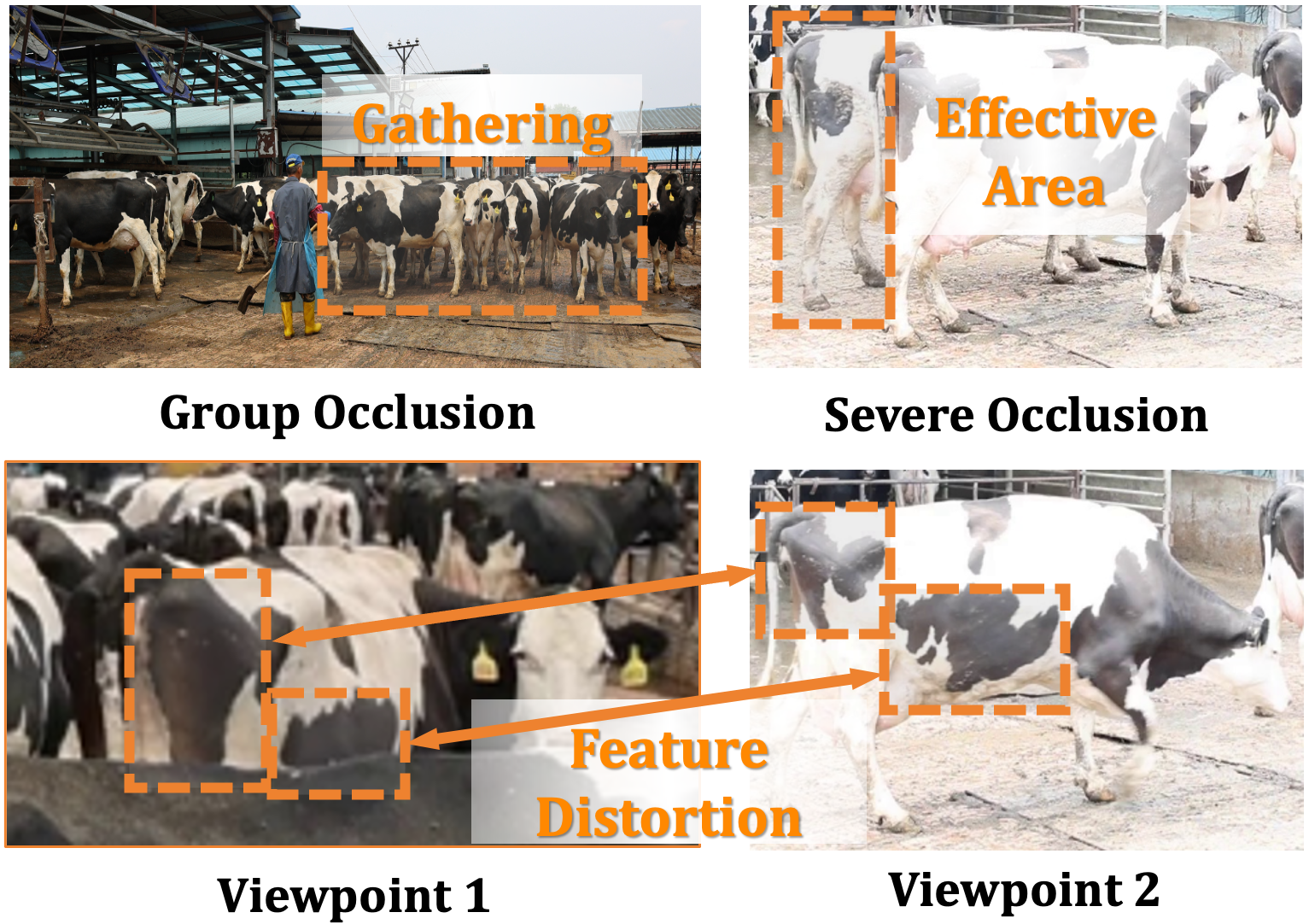}
	\caption{Representative samples from the MVCow dataset illustrating herd occlusion, severe individual occlusion, and cross-view appearance distortion.}
	\label{fig:mvcow_samples}
\end{figure}

\begin{figure}
	\centering
	\includegraphics[width=0.95\linewidth]{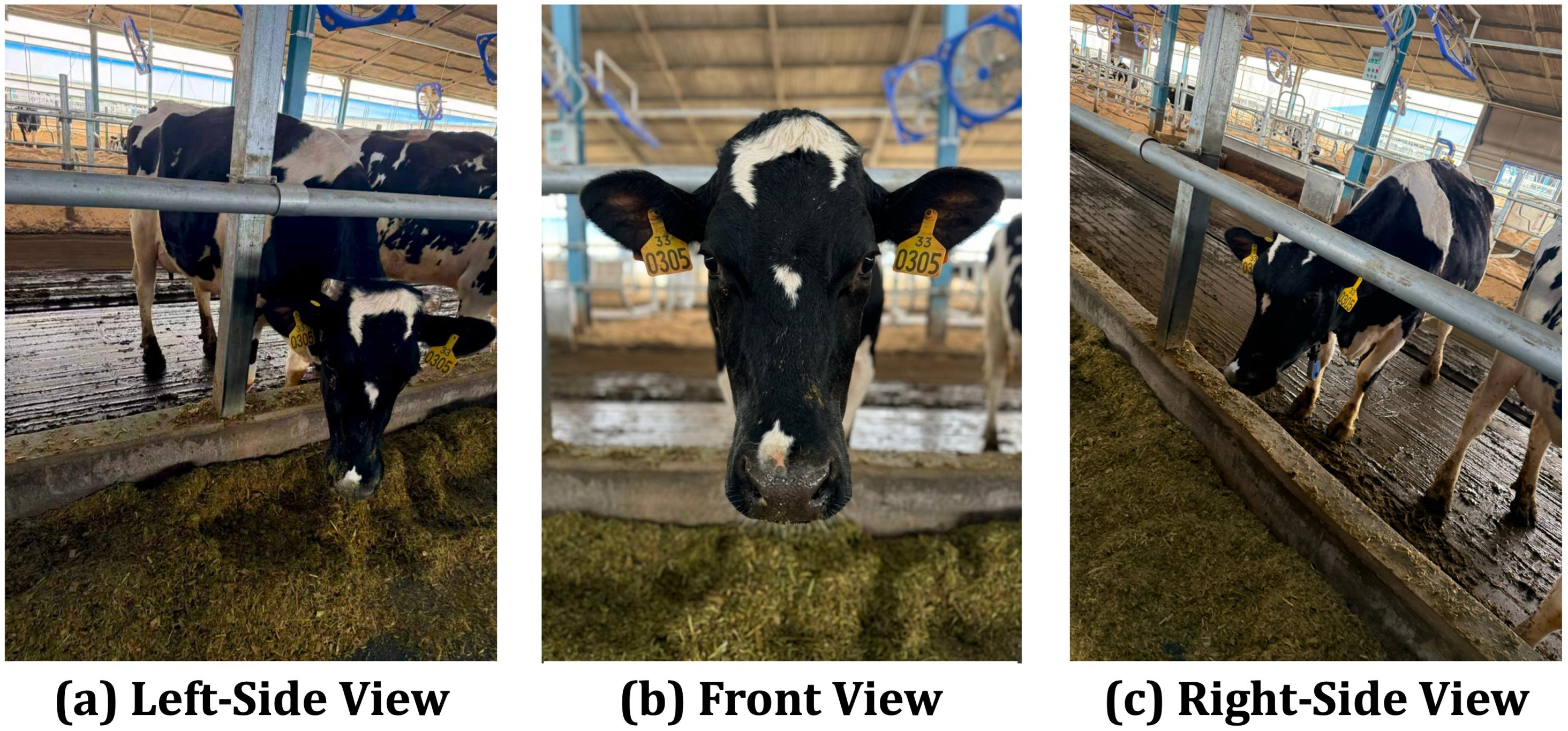}
	\caption{Multi-view images of the same individual cow, highlighting the visual variations across different camera angles.}
	\label{fig:multi_view_sample}
\end{figure}

\textbf{MVCow Dataset:} Collected using two synchronized, non-overlapping cameras at the junction of a milking channel and parlor, MVCow contains 12 video segments and 10,184 manually annotated frames covering 148 uniquely identified cattle. It provides comprehensive annotations for object detection, multi-object tracking, and cross-camera ReID, featuring severe occlusions (Fig.~\ref{fig:mvcow_samples}) and drastic cross-view appearance distortions (Fig.~\ref{fig:multi_view_sample}). The task-specific data split configurations are summarized in Table~\ref{tab:mvcow_splits}.

\begin{figure*}
	\centering
	\includegraphics[width=0.95\linewidth]{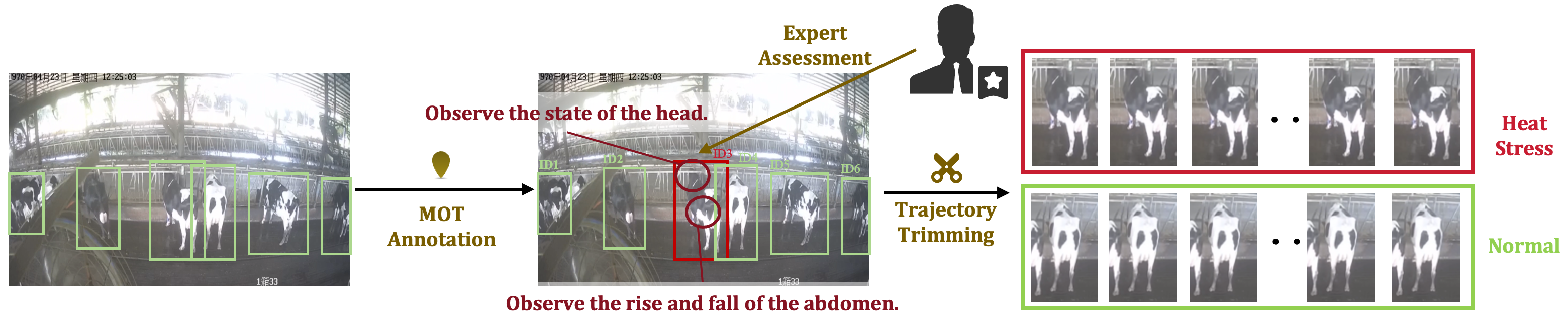}
	\caption{The annotation workflow for the HSCow heat stress dataset, including multi-object tracking annotation, tracklet clipping, and expert evaluation.}
	\label{fig:hscow_annotation}
\end{figure*}

\textbf{HSCow Dataset:} Collected in the cooling zone of a commercial barn during summer, HSCow focuses on fine-grained physiological behaviors. It comprises 1,131 individual tracklet segments (Table~\ref{tab:hscow_statistics}), categorized into 748 normal and 383 heat stress samples based on expert veterinary assessments (the detailed annotation workflow is depicted in Fig.~\ref{fig:hscow_annotation}).

\begin{table*}
	\centering
	\caption{Data collection information for the constructed benchmarks.}
	\label{tab:data_collection}
	\begin{tabular}{@{}llp{4cm}lp{1.5cm}p{4.5cm}@{}}
		\toprule
		\textbf{Dataset} & \textbf{Farm Location} & \textbf{Area} & \textbf{Time} & \textbf{Videos} & \textbf{Purpose} \\
		\midrule
		MVCow & Yanqing, Beijing & Intersection of milking channel and parlor & Apr 2025 & 12 Clips & Train and evaluate object detection, ReID, and MOT models \\
		\addlinespace
		HSCow & Shandong & Barn cooling zone & Jun-Sep 2025 & 6 Clips & Train and evaluate heat stress behavior recognition models \\
		\bottomrule
	\end{tabular}
\end{table*}

\begin{table*}
	\centering
	\caption{Task-specific data split configurations for the MVCow dataset.}
	\label{tab:mvcow_splits}
	\begin{tabular}{@{}llccc@{}}
		\toprule
		\textbf{Task} & \textbf{Data Format} & \textbf{Split Method} & \textbf{Training Set} & \textbf{Testing Set} \\
		\midrule
		Object Detection & YOLO & Frame & 7,128 Frames & 3,056 Frames \\
		Multi-Object Tracking & MOT17 & Video Clip & --- & 12 Clips \\
		Re-Identification & Market-1501 & Identity & 115 IDs / 54,230 Images & \begin{tabular}[c]{@{}c@{}}Query: 33 IDs / 103 Images\\ Gallery: 33 IDs / 24,223 Images\end{tabular} \\
		\bottomrule
	\end{tabular}
\end{table*}

\begin{table}
	\centering
	\caption{Statistical overview of the HSCow behavior dataset.}
	\label{tab:hscow_statistics}
	\begin{tabular}{@{}lcc@{}}
		\toprule
		\textbf{Category} & \textbf{Number of Clips} & \textbf{Avg. Duration (s)} \\
		\midrule
		Normal & 748 & 15.17 \\
		Heat Stress & 383 & 13.78 \\
		\midrule
		\textbf{Total} & \textbf{1131} & \textbf{14.70} \\
		\bottomrule
	\end{tabular}
\end{table}

\subsection{Framework Overview}
The primary objective of the CowCare framework is to bridge the gap between raw multi-camera video streams and individual-level heat stress diagnostics in uncalibrated commercial barns. To achieve this, we decouple the task into two sequential stages: establishing reliable spatiotemporal individual tracklets, and extracting fine-grained behavioral semantics from these tracklets. 

\begin{figure}
	\centering
	\includegraphics[width=0.95\linewidth]{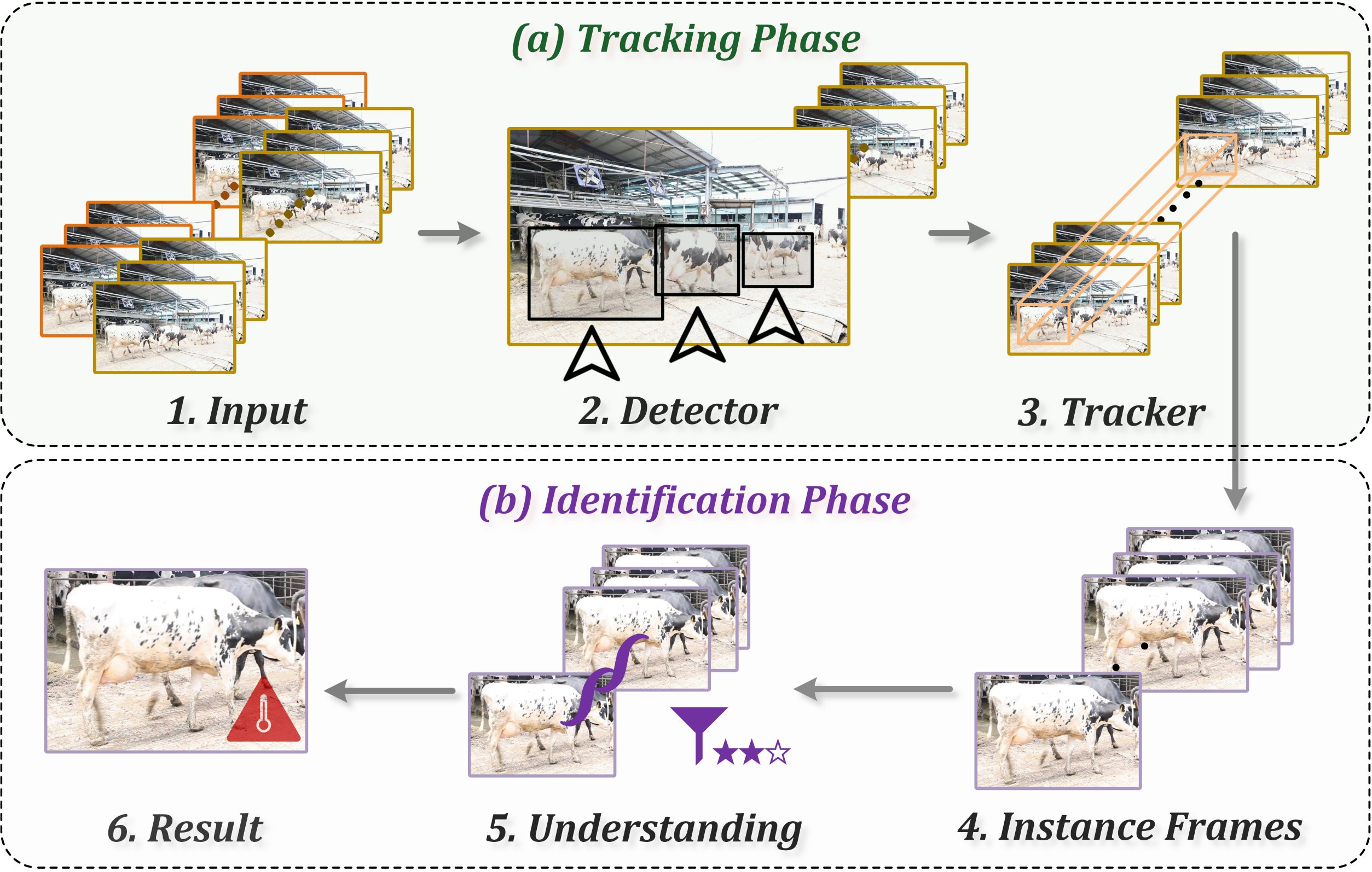}
	\caption{The overall architecture of the CowCare framework.}
	\label{fig:overall_pipeline}
\end{figure}

As illustrated in Fig.~\ref{fig:overall_pipeline}, CowCare operates as a unified two-stage pipeline. In the first stage, video streams from non-overlapping cameras are fed into a YOLO-based detector and our proposed CowTrack pipeline. CowTrack utilizes the View-Decoupled Aggregation Network (VDA-Net) alongside a causal online self-supervised strategy to associate identities and generate stable, long-term tracklets. In the second stage, these localized individual tracklets are processed by TSDF-Net. By filtering out physical camera jitter and injecting frozen diffusion semantic priors, TSDF-Net classifies the subtle spatiotemporal patterns of heat stress, yielding the final diagnostic predictions.

\subsection{Viewpoint-Robust Tracking (CowTrack)}
\label{sec:cowtrack}

\begin{figure*}
	\centering
	\includegraphics[width=0.95\linewidth]{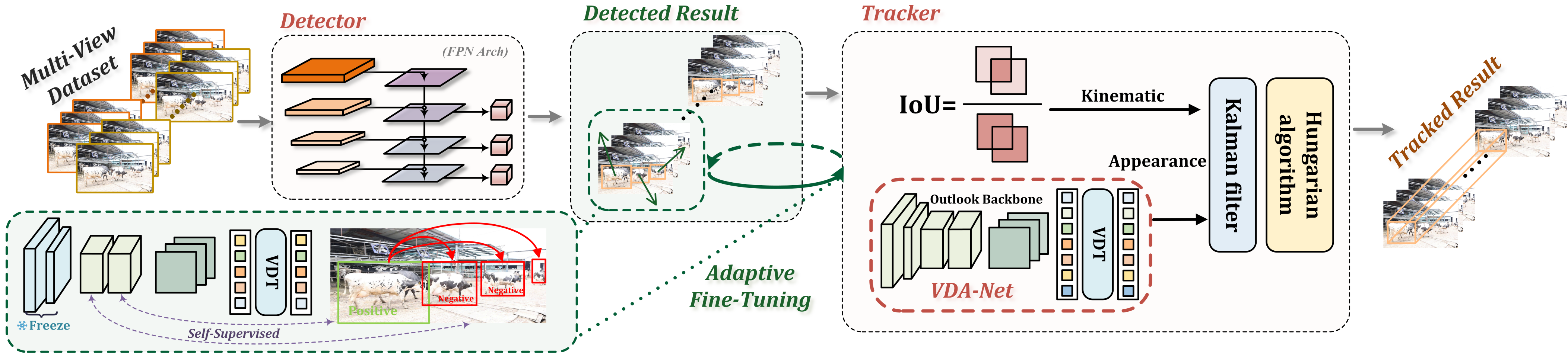}
	\caption{The overall architecture of the CowTrack framework.}
	\label{fig:cowtrack_framework}
\end{figure*}

\begin{figure*}
	\centering
	\includegraphics[width=0.95\linewidth]{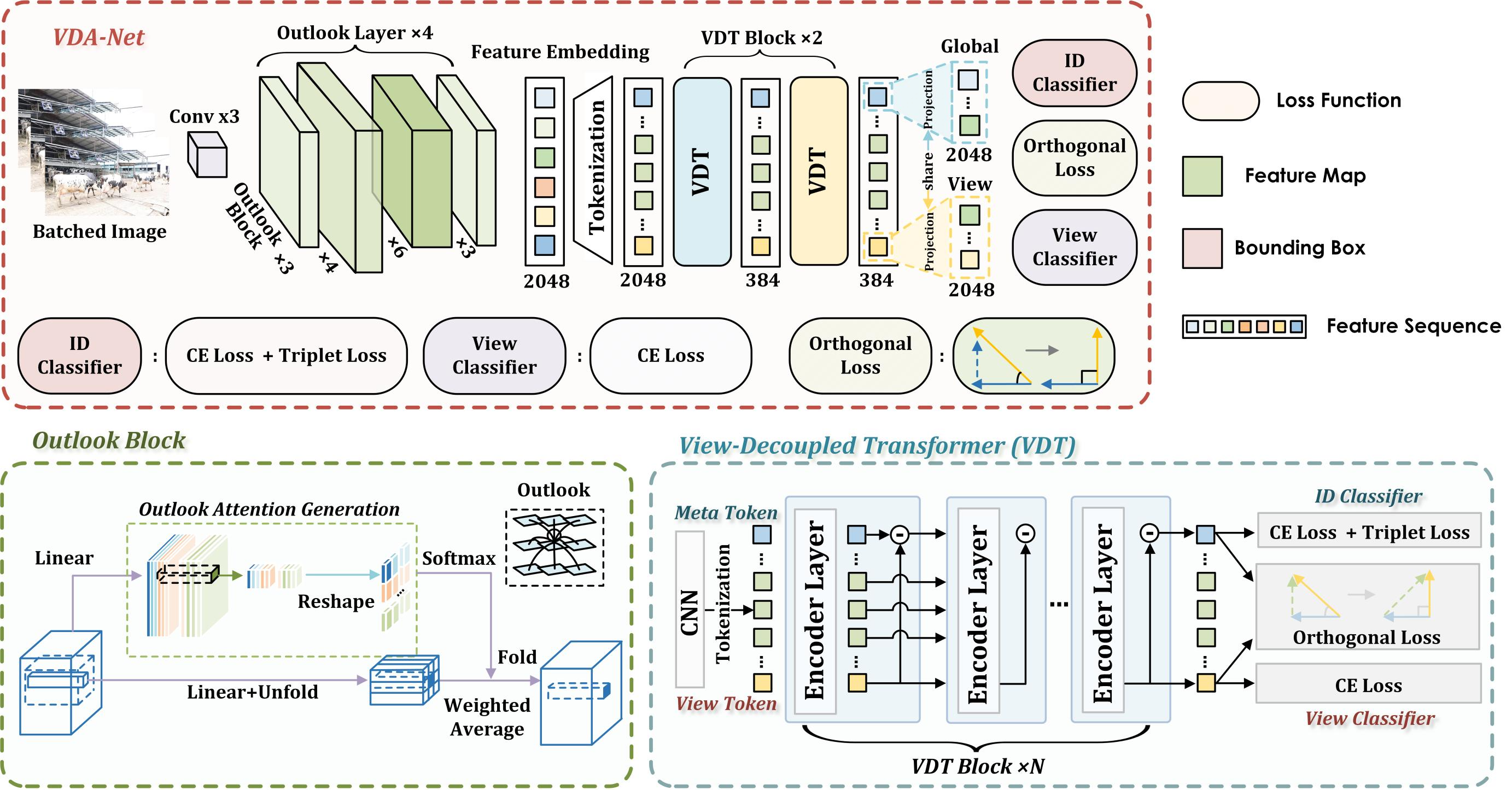}
	\caption{Architecture and core components of the View-Decoupled Aggregation Network (VDA-Net). \textbf{Top:} The overall VDA-Net pipeline, illustrating the backbone integrated with Outlook attention layers and the Viewpoint Decoupling Transformer (VDT). \textbf{Bottom left:} The structure of the Outlook block, utilized for localized texture aggregation under severe occlusions. \textbf{Bottom right:} Detailed structure of the VDT block, showcasing the subtractive mechanism for isolating view-invariant identity embeddings.}
	\label{fig:vdanet}
\end{figure*}

Establishing continuous trajectories across multiple cameras is a prerequisite for individual-level behavior analysis. However, dense herds cause severe partial occlusions, and non-overlapping camera networks induce substantial viewpoint and scale distortions. As illustrated in Fig.~\ref{fig:cowtrack_framework}, CowTrack integrates YOLO11 for detection and our proposed VDA-Net (detailed in Fig.~\ref{fig:vdanet}) as the core appearance extractor. VDA-Net departs from the conventional global-feature paradigm by adopting localized aggregation and explicit viewpoint decoupling. Motivated by the complementary roles of motion and appearance in tracking-by-detection, as exemplified by BoostTrack~\citep{stanojevic2024boosttrack++}, CowTrack combines kinematic cues with the occlusion- and viewpoint-robust embeddings learned by VDA-Net for within-camera temporal association.

\subsubsection{Localized Aggregation for Severe Occlusion}
Under dense overlap, global features relying on the full body silhouette become highly unstable. Our design is inspired by the localized spatial reweighting principle of Outlook attention~\citep{yuan2022volo}, which we adapt to multi-scale cattle identity encoding. Specifically, VDA-Net places Outlook blocks across multiple feature-pyramid levels and learns position-specific weights for the visible coat patches at each scale, rather than relying on a holistic body representation. This task-oriented adaptation preserves fine-grained identity evidence even when only the head or flank is observable.

\subsubsection{Viewpoint Decoupling Transformer (VDT)}
Even with preserved local textures, an animal's appearance shifts significantly when transitioning between camera fields of view. Prior view-decoupled representation learning motivates modeling identity and viewpoint as distinct factors~\citep{zhang2024view}. Inspired by this principle, we integrate VDT (Fig.~\ref{fig:vdanet}) for non-overlapping cattle cameras by coupling a camera-supervised view token with hierarchical subtraction. The patch tokens $P^j$ are derived from the multi-scale feature maps. The meta token $t_m^j$ is trained with the global cattle identity shared across cameras, whereas the view token $t_v^j$ is supervised by the camera-stream label. 

At block $j$, a Transformer encoder~\citep{dosovitskiy2020image} first updates the meta token, patch tokens, and view token:
\begin{equation}
\begin{bmatrix}
\bar{t}_m^{j+1} \\ \bar{P}^{j+1} \\ \bar{t}_v^{j+1}
\end{bmatrix}
=F^j\!\left(
\begin{bmatrix}
t_m^j \\ P^j \\ t_v^j
\end{bmatrix}
\right).
\label{eq:vdt_attention}
\end{equation}

Through a hierarchical subtractive mechanism, the view\-point-dependent components are explicitly disentangled from the meta-token:
\begin{equation}
t_m^{j+1}=\bar{t}_m^{j+1}-\bar{t}_v^{j+1},\quad
t_v^{j+1}=\bar{t}_v^{j+1},\quad
P^{j+1}=\bar{P}^{j+1}.
\label{eq:vdt_subtraction}
\end{equation}
This isolates an intrinsic, view-invariant identity embedding specifically optimized for cross-camera association.

\subsubsection{Offline Training Objectives}
The offline objectives enforce the representation roles introduced above. Let $B$ be a supervised mini-batch, $C$ the number of training identities, and $V$ the number of camera views. For sample $i$, the corresponding cross-entropy objectives for identity and view are:
\begin{align}
\mathcal{L}_{\mathrm{ID}} &=-\frac{1}{|B|}\sum_{i=1}^{|B|}\sum_{c=1}^{C}y_{ic}\log p_{ic}, \label{eq:id_loss}\\
\mathcal{L}_{\mathrm{View}} &=-\frac{1}{|B|}\sum_{i=1}^{|B|}\sum_{r=1}^{V}v_{ir}\log q_{ir}. \label{eq:view_loss}
\end{align}

Identity discrimination is additionally optimized with a triplet loss \citep{schroff2015facenet}:
\begin{equation}
\mathcal{L}_{\mathrm{Tri}}=
\frac{1}{|B|}\sum_{i=1}^{|B|}
\left[
\lVert f_{m,i}-f_{m,i}^{+}\rVert_2^2
-\lVert f_{m,i}-f_{m,i}^{-}\rVert_2^2+\alpha
\right]_+.
\label{eq:offline_triplet}
\end{equation}

To reduce redundancy between identity and view embeddings, we minimize their absolute cosine similarity:
\begin{equation}
\mathcal{L}_{\mathrm{Orth}}=
\frac{1}{|B|}\sum_{i=1}^{|B|}
\left|
\frac{f_{m,i}^{\top}f_{v,i}}
{\lVert f_{m,i}\rVert_2\lVert f_{v,i}\rVert_2}
\right|.
\label{eq:orthogonal_loss}
\end{equation}

The complete offline objective is defined as $\mathcal{L}_{\mathrm{offline}} = \mathcal{L}_{\mathrm{ID}} + \lambda_{\mathrm{tri}}\mathcal{L}_{\mathrm{Tri}} + \lambda_{\mathrm{view}}\mathcal{L}_{\mathrm{View}} + \lambda_{\mathrm{orth}}\mathcal{L}_{\mathrm{Orth}}$.

\subsubsection{Causal Online Self-Supervised Adaptation}
\begin{figure*}
	\centering
	\includegraphics[width=0.8\linewidth]{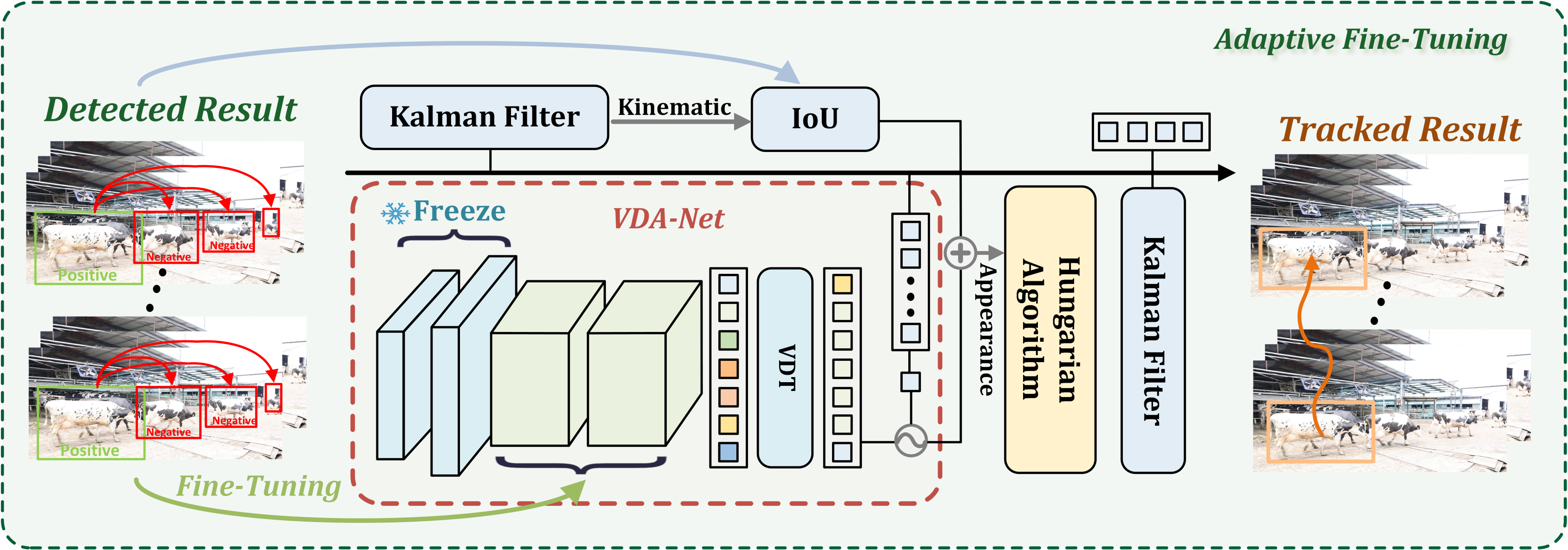}
	\caption{Flowchart of the causal online self-supervised fine-tuning process for dynamic scene adaptation.}
	\label{fig:adaptive_finetune}
\end{figure*}

An offline-trained representation suffers from domain shift when deployed in a dynamic barn environment due to shifting illumination and novel herd compositions. Therefore, CowTrack incorporates a causal online self-supervised adaptation strategy, as shown in Fig.~\ref{fig:adaptive_finetune}. Leveraging the spatiotemporal continuity of short-term tracklets, the tracker generates pseudo-labels on the fly to continuously update VDA-Net. 

At frame $t$, the sampler constructs an $N$-way, $K$-shot mini-batch from the available pseudo-identities $B_t$. Given this causally formed batch, established metric-learning objectives provide stable optimization signals. The distinguishing element is that all supervision is constructed exclusively from identities available at or before frame $t$. Center loss~\citep{wen2016discriminative} promotes compact features within each pseudo-identity $c_i$:
\begin{equation}
\mathcal{L}_{\mathrm{Center}}=
\frac{1}{NK}\sum_{i=1}^{N}\sum_{k=1}^{K}
\lVert f_{i,k}-c_i\rVert_2^2,
\label{eq:center_loss}
\end{equation}
For normalized embeddings, supervised contrastive loss \citep{khosla2020supervised}, which optimizes intra-class compactness and inter-class dispersion~\citep{wang2021understanding}, is applied as:
\begin{equation}
\mathcal{L}_{\mathrm{Con}}=
-\frac{1}{|B_t|}\sum_{a\in B_t}
\frac{1}{|P(a)|}\sum_{p\in P(a)}
\log
\frac{\exp(f_a^{\top}f_p/\tau)}
{\sum_{b\in B_t\setminus\{a\}}\exp(f_a^{\top}f_b/\tau)}.
\label{eq:contrastive_loss}
\end{equation}

The total online adaptation loss is $\mathcal{L}_{\mathrm{online}} = w_{\mathrm{tri}}\mathcal{L}_{\mathrm{Tri}} + w_{\mathrm{center}}\mathcal{L}_{\mathrm{Center}} + w_{\mathrm{con}}\mathcal{L}_{\mathrm{Con}}$. All updates use only observations available up to frame $t$, preserving causality at inference.

\subsection{Noise-Resistant and Semantics-Enhanced Behavior Recognition (TSDF-Net)}
\label{sec:tsdfnet}
\begin{figure*}
	\centering
	\includegraphics[width=0.8\linewidth]{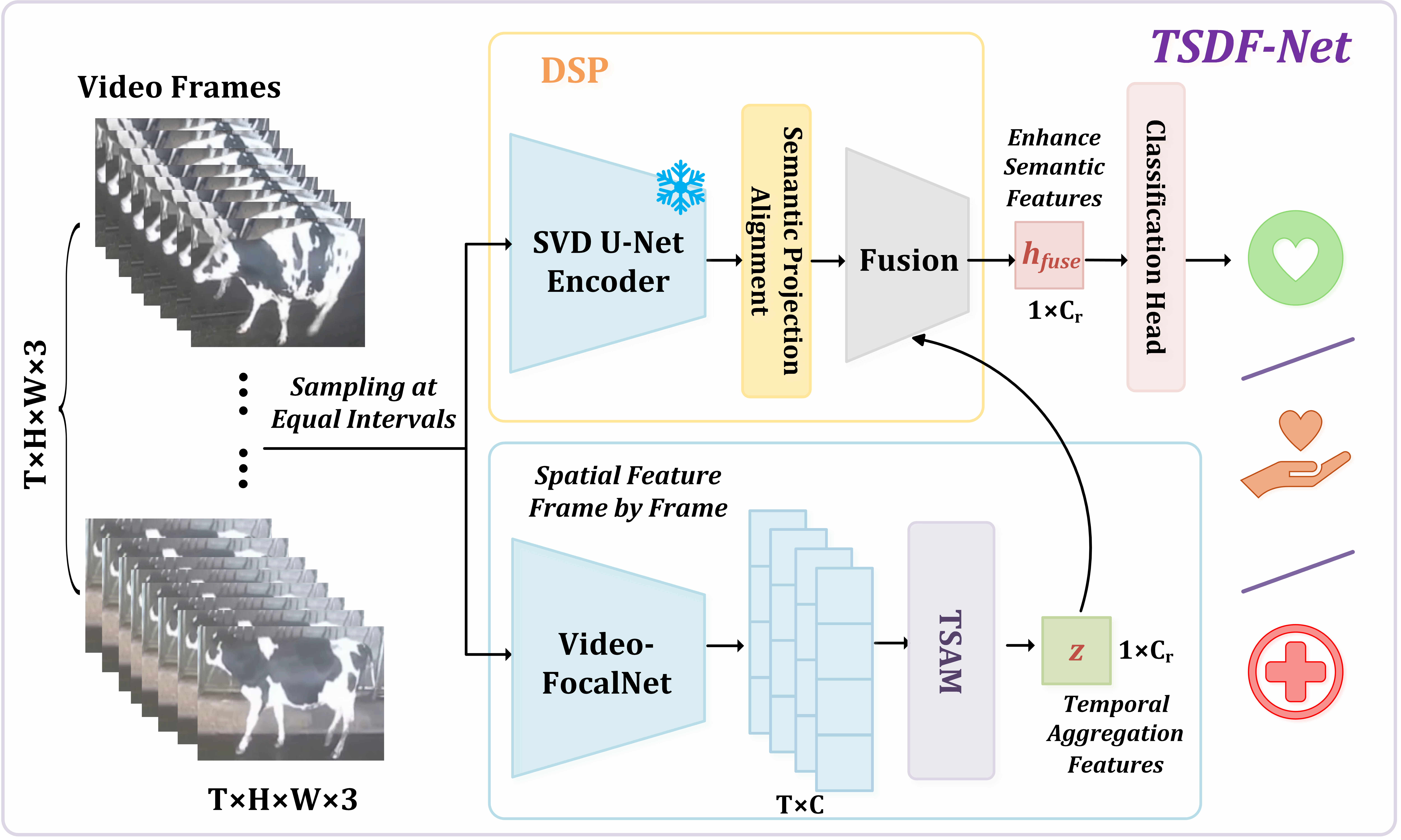}
	\caption{The overall architecture of TSDF-Net, highlighting the Temporal Stabilization and Aggregation Module (TSAM) and the Diffusion Semantic Prior (DSP).}
	\label{fig:tsdfnet}
\end{figure*}

Once stable individual tracklets are extracted, the subsequent challenge is decoding the subtle spatiotemporal signals of heat stress (e.g., accelerated respiration). However, real-world barn videos are inherently noisy due to the physical vibrations of cooling fans, and acquiring large-scale, veterinary-grade behavioral datasets is prohibitively expensive. To address the dual challenges of environmental jitter and data scarcity, we propose TSDF-Net, whose overall architecture is illustrated in Fig.~\ref{fig:tsdfnet}.

\subsubsection{Temporal Stabilization and Aggregation Module (TSAM)}

\begin{figure*}
	\centering
	\includegraphics[width=0.95\linewidth]{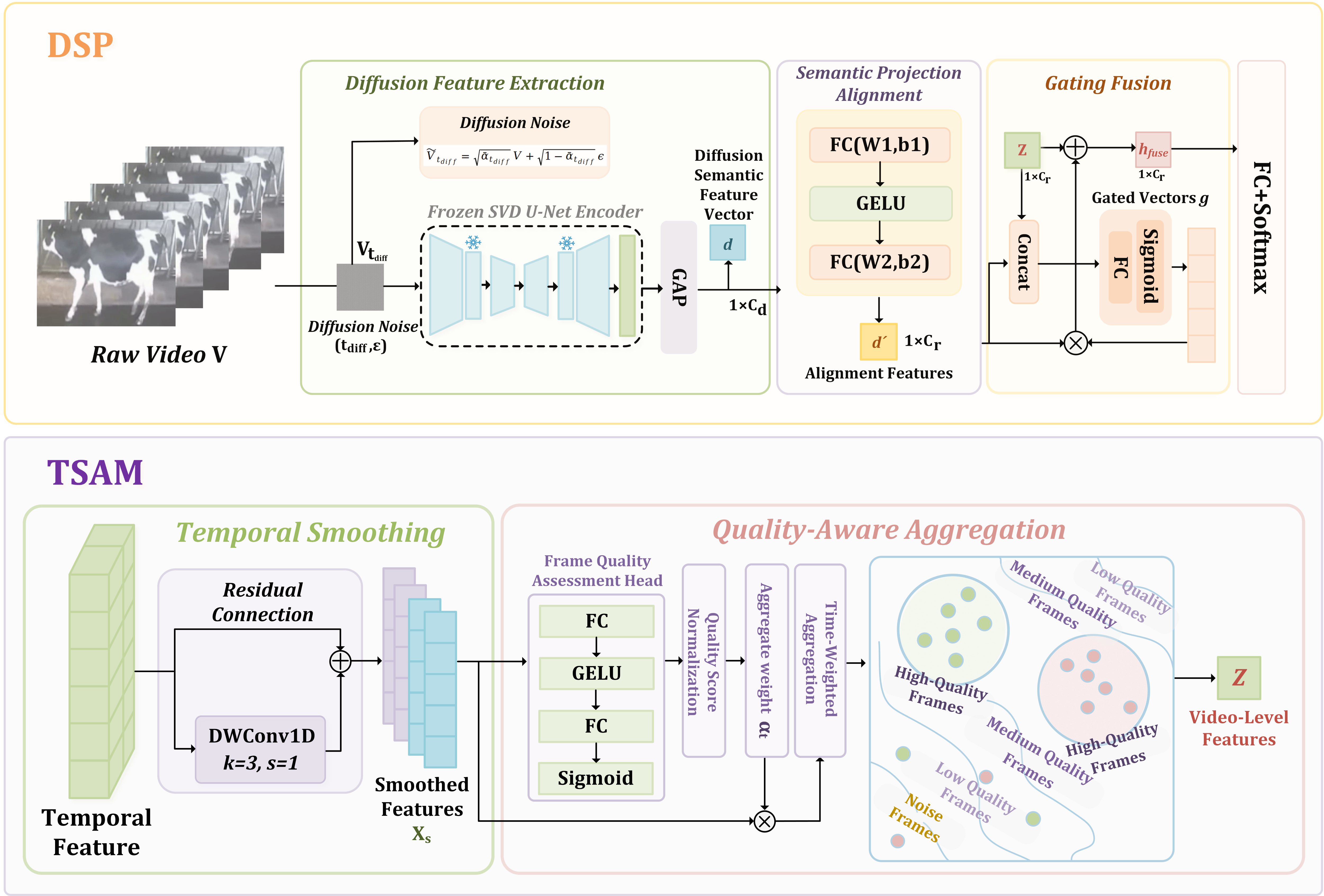}
	\caption{Schematic of the Temporal Stabilization and Aggregation Module (TSAM) and the Diffusion Semantic Prior (DSP). TSAM performs temporal smoothing through 1D depthwise convolution and quality-aware gating, whereas DSP injects generalized motion semantics through a frozen diffusion encoder.}
	\label{fig:tsp_tsam}
\end{figure*}

Direct temporal aggregation in unstable videos confounds subtle physiological rhythms with high-frequency camera vibrations. To mitigate this, we introduce TSAM (shown in Fig.~\ref{fig:tsp_tsam}) to explicitly filter out physical noise. A learnable 1D depthwise convolution is applied along the temporal axis of the frame-level feature sequence to perform low-pass filtering. This preserves low-frequency respiratory rhythms while dampening high-frequency jitter. Following this temporal smoothing, a quality-aware gating network evaluates the reliability of each frame, assigning lower aggregation weights to blurred or jitter-corrupted frames. This ensures that the final spatiotemporal representation is dominated by the most stable and informative visual cues.

\subsubsection{Diffusion Semantic Prior (DSP)}
Pre-trained video diffusion models encode generalized motion structure and scene semantics within their U-Net representations~\citep{ho2022video, blattmann2023stable}. Motivated by this property, we design DSP to transfer these priors to fine-grained livestock behavior recognition. Rather than fine-tuning the diffusion model or treating it as an additional prediction branch, DSP uses its frozen encoder as an external semantic prior and learns how to align that prior with the noise-stabilized representation produced by TSAM.

In DSP (Fig.~\ref{fig:tsp_tsam}), slight Gaussian noise is added to the tracklet sequence before it is processed by a frozen SVD U-Net encoder~\citep{blattmann2023stable}. The resulting diffusion features are projected into the recognition space and fused with the TSAM output through a learnable gate. This task-conditioned coupling allows the generalized motion prior to complement, rather than replace, the behavior-specific temporal representation, improving sensitivity to subtle heat-stress dynamics while limiting the number of trainable parameters on the small dataset.

\subsection{Evaluation Metrics}
We emphasize task-specific core metrics. For completeness and direct comparison with prior work, the result tables also report Rank-5, macro-precision, and macro-recall as secondary metrics.

For ReID, Rank-1 accuracy measures whether the correct identity is retrieved first, and mean average precision (mAP) summarizes ranking quality over all gallery samples~\citep{zheng2015scalable}:

\begin{equation}
\mathrm{Rank}\text{-}1=
\frac{1}{N_q}\sum_{q=1}^{N_q}\mathbf{1}(r_q=1),
\label{eq:rank_k}
\end{equation}
\begin{equation}
\mathrm{AP}_q=
\frac{1}{M_q}\sum_{r=1}^{N_g}P_q(r)\,\mathrm{rel}_q(r),
\quad
\mathrm{mAP}=\frac{1}{N_q}\sum_{q=1}^{N_q}\mathrm{AP}_q.
\label{eq:map}
\end{equation}
Here, $N_q$ and $N_g$ are the numbers of query and gallery samples, respectively, $r_q$ is the rank of the first correct match for query $q$, $M_q$ is the number of relevant gallery samples, $P_q(r)$ is the precision among the top-$r$ retrievals, $\mathrm{rel}_q(r)$ equals 1 if the item at rank $r$ is relevant to query $q$ and 0 otherwise, and $\mathbf{1}(\cdot)$ is the indicator function.

For tracking, HOTA jointly measures detection and association quality~\citep{luiten2021hota}, MOTA summarizes false negatives, false positives, and identity switches relative to the total number of ground-truth objects~\citep{bernardin2008clear}, and IDF1 measures identity preservation~\citep{ristani2016performance}:
\begin{align}
\mathrm{DetA}_{\alpha}
&=\frac{|\mathrm{TP}_{\alpha}|}
{|\mathrm{TP}_{\alpha}|+|\mathrm{FP}_{\alpha}|+|\mathrm{FN}_{\alpha}|},
\label{eq:deta}\\
\mathrm{HOTA}
&=\frac{1}{19}\sum_{\alpha\in\mathcal{A}}
\sqrt{\mathrm{DetA}_{\alpha}\,\mathrm{AssA}_{\alpha}}.
\label{eq:hota}
\end{align}
\begin{equation}
\mathrm{MOTA}=1-
\frac{\sum_{t=1}^{T}(\mathrm{FN}_t+\mathrm{FP}_t+\mathrm{IDSW}_t)}
{\sum_{t=1}^{T}\mathrm{GT}_t},
\label{eq:mota}
\end{equation}
\begin{equation}
\mathrm{IDF1}=
\frac{2\,\mathrm{IDTP}}
{2\,\mathrm{IDTP}+\mathrm{IDFP}+\mathrm{IDFN}},
\label{eq:idf1}
\end{equation}
At localization threshold $\alpha$, $\mathrm{TP}_{\alpha}$, $\mathrm{FP}_{\alpha}$, and $\mathrm{FN}_{\alpha}$ denote true-positive, false-positive, and false-negative detections, $\mathrm{DetA}_{\alpha}$ is detection accuracy, $\mathrm{AssA}_{\alpha}$ is the mean association accuracy of matched trajectories, and $\mathcal{A}=\{0.05,0.10,\ldots,0.95\}$ is the set of 19 evaluation thresholds. For MOTA, $T$ is the number of frames, while $\mathrm{FN}_t$, $\mathrm{FP}_t$, $\mathrm{IDSW}_t$, and $\mathrm{GT}_t$ are the numbers of false negatives, false positives, identity switches, and ground-truth objects in frame $t$, respectively. IDTP, IDFP, and IDFN are correctly identified, incorrectly identified, and missed identity detections, respectively.

For heat-stress recognition, Top-1 accuracy and macro-F1 are the primary metrics:
\begin{equation}
\mathrm{Top\text{-}1}
 =\frac{1}{N}\sum_{i=1}^{N}\mathbf{1}(\hat{y}_i=y_i).
\label{eq:top1}
\end{equation}
\begin{equation}
\mathrm{F1}_{\mathrm{macro}}
 =\frac{1}{C}\sum_{c=1}^{C}
\frac{2\,\mathrm{P}_c\mathrm{R}_c}{\mathrm{P}_c+\mathrm{R}_c}.
\label{eq:recognition_metrics}
\end{equation}
Here, $N$ is the number of test clips, $y_i$ and $\hat{y}_i$ are the ground-truth and predicted labels of clip $i$, $C$ is the number of classes, $\mathrm{P}_c=\mathrm{TP}_c/(\mathrm{TP}_c+\mathrm{FP}_c)$ and $\mathrm{R}_c=\mathrm{TP}_c/(\mathrm{TP}_c+\mathrm{FN}_c)$ are the precision and recall for class $c$, and $\mathrm{TP}_c$, $\mathrm{FP}_c$, and $\mathrm{FN}_c$ are the corresponding true-positive, false-positive, and false-negative counts.

\section{Results and Discussion}
\label{sec:experiments}

\subsection{Implementation Details}
CowTrack utilized YOLO11~\citep{khanam2024yolov11}, which benefits from advanced bounding box regression and dense detection objectives~\citep{zheng2020distance, lin2017focal}, for robust object detection. VDA-Net was initialized with ImageNet weights and trained using SGD for 150 epochs with an initial learning rate of $8 \times 10^{-3}$. During tracking, the online self-supervised adaptation was triggered periodically, utilizing an $N$-way $K$-shot sampling strategy ($N=8, K=4$) exclusively from the tracking buffer. TSDF-Net adopted Video-FocalNet~\citep{wasim2023video} as its spatial backbone. Unless otherwise specified, TSDF-Net used $T=8$ input frames with a frame interval of $\Delta t=8$. The network was optimized using AdamW for 30 epochs with an initial learning rate of $5 \times 10^{-4}$. The SVD U-Net encoder in the DSP was kept strictly frozen during all training phases.

\subsection{Evaluation of Identity Preservation (CowTrack)}

\begin{table*}
\centering
\caption{Cross-camera ReID performance on MVCow (\%). Best and second-best results are shown in bold and underlined, respectively. $\uparrow$ indicates higher values are better.}
\label{tab:reid_comparison}
\begin{tabular}{lccc}
\toprule
\textbf{Method} & \textbf{Rank-1 $\uparrow$} & \textbf{Rank-5 $\uparrow$} & \textbf{mAP $\uparrow$} \\
\midrule
CLIP-ReID~\citep{li2023clip} & 97.36 & 99.84 & 82.45 \\
DGNet~\citep{zheng2019joint} & 92.89 & 97.81 & 81.79 \\
HOReID~\citep{wang2021horeid} & 61.28 & 62.89 & 62.67 \\
OSNet~\citep{zhou2019omni} & \underline{98.90} & \underline{99.90} & 99.60 \\
PCB~\citep{sun2018beyond} & 61.79 & 63.04 & 63.07 \\
PersonViT~\citep{hu2025personvit} & 94.90 &	99.10 & 93.20\\
ResNet-IBN~\citep{pan2018two} & 97.72 & 99.71 & \underline{99.65} \\
ResNeSt~\citep{zhang2022resnest} & 96.03 & 96.22 & 96.69 \\
TransReID~\citep{he2021transreid} & 98.70 & \underline{99.90} & 99.50 \\
\midrule
\textbf{VDA-Net (Ours)} & \textbf{99.83} & \textbf{99.98} & \textbf{99.67} \\
\bottomrule
\end{tabular}
\end{table*}

As shown in Table~\ref{tab:reid_comparison}, VDA-Net achieves top performance across all ReID metrics, reaching 99.83\% in Rank-1 and 99.67\% in mAP. Notably, while Rank-1 accuracy approaches its performance ceiling, VDA-Net maintains a significant advantage in mAP. This indicates that the improvement stems from an overall optimization of retrieval ranking quality rather than merely first-position matching. Among competitive baselines, OSNet and TransReID demonstrate strong capabilities. However, VDA-Net still outperforms them, suggesting that our VDA-Net backbone exhibits stronger robustness against occlusion and viewpoint variations. Conversely, while CLIP-ReID boasts high Rank-k scores, its mAP drops to 82.45\%, revealing that its pre-trained features are biased towards coarse-grained class-level semantics and struggle to capture subtle individual intra-class differences.

\begin{table*}
	\centering
	\caption{Global multi-object tracking performance on MVCow (\%). Best and second-best results are shown in bold and underlined, respectively. $\uparrow$ indicates higher values are better.}
	\label{tab:tracking_comparison}
	\begin{tabular}{lccc}
		\toprule
		\textbf{Method} & \textbf{HOTA $\uparrow$} & \textbf{MOTA $\uparrow$} & \textbf{IDF1 $\uparrow$} \\
		\midrule
		BoostTrack~\citep{stanojevic2024boosttrack++} & 77.49 & 82.76 & 81.40 \\
		ByteTrack~\citep{zhang2022bytetrack} & 75.18 & 79.47 & \underline{82.09} \\
		BoT-SORT~\citep{aharon2022bot} & 68.87 & 79.04 & 72.11 \\
		DeepSORT~\citep{wojke2017simple} & 75.07 & 80.03 & 81.30 \\
		Deep OC-SORT~\citep{maggiolino2023deep} & 75.29 & 80.00 & 81.58 \\
		Depth-OC-SORT~\citep{zhu2026depth} & 74.99 & 75.77 & 80.13 \\
		HybridSORT~\citep{yang2024hybrid} & \underline{78.53} & \underline{89.24} & 81.96 \\
		OC-SORT~\citep{cao2023observation} & 75.15 & 79.79 & 81.02 \\
		StrongSORT~\citep{du2023strongsort} & 74.86 & 80.00 & 81.01 \\
		\midrule
		\textbf{CowTrack (Ours)} & \textbf{83.48} & \textbf{89.50} & \textbf{87.72} \\
		\bottomrule
	\end{tabular}
\end{table*}

Table~\ref{tab:tracking_comparison} compares the global tracking performance of CowTrack with existing methods. CowTrack obtains 83.48\% HOTA, 89.50\% MOTA, and 87.72\% IDF1, improving upon BoostTrack by 5.99 percentage points (pp), 6.74 pp, and 6.32 pp, respectively. OC-SORT and ByteTrack primarily associate detections using motion cues, which become unreliable during prolonged occlusion and can produce fragmented trajectories. HybridSORT achieves a MOTA of 89.24\%, close to that of CowTrack, but its HOTA and IDF1 are lower by 4.95 pp and 5.76 pp, respectively. This comparison shows that a high MOTA does not necessarily imply consistent identity association. CowTrack addresses this limitation through online self-supervised adaptation, which updates the appearance representation as scene conditions change and improves identity preservation in crowded farm environments.

\subsection{Evaluation of Heat Stress Recognition (TSDF-Net)}

\begin{table*}
	\centering
	\caption{Heat stress behavior recognition performance on the HSCow dataset (\%). The best results are shown in bold, while the second-best results are underlined. $\uparrow$ indicates that higher values are better.}
	\label{tab:heat_stress_results}
	
	\begin{tabular}{lcccc}
		\toprule
		\textbf{Model} & \textbf{Top-1 $\uparrow$} & \textbf{Precision $\uparrow$} & \textbf{Recall $\uparrow$} & \textbf{F1 $\uparrow$} \\
		\midrule
		ActionDiff~\citep{guimaraes2026diffusion} & 76.32 & 78.30 & 76.08 & 75.78 \\
		SlowOnly-R50~\citep{feichtenhofer2019slowfast} & 78.07 & 82.36 & 78.39 & 77.44 \\
		SlowFast~\citep{feichtenhofer2019slowfast}  & 70.18 & 81.11 & 70.69 & 67.62 \\
		TSN~\citep{wang2016temporal}   & 78.07 & 78.51 & 77.96 & 77.93 \\
		TSM~\citep{lin2019tsm} & 76.32 & 76.97 & 76.45 & 76.23 \\
		TimeSformer~\citep{bertasius2021space}  & 70.18 & 70.55 & 70.04 & 69.94 \\
		Video Swin~\citep{liu2022video}  & 59.65 & 73.77 & 58.96 & 51.80 \\
		VideoMAE~\citep{tong2022videomae}  & 71.93 & 72.02 & 71.86 & 71.85 \\
		Video-FocalNet~\citep{wasim2023video} 
		& \underline{82.46} 
		& \underline{83.06} 
		& \underline{82.57} 
		& \underline{82.41} \\
		\midrule
		\textbf{TSDF-Net (Ours)} 
		& \textbf{87.72} 
		& \textbf{89.35} 
		& \textbf{87.90} 
		& \textbf{87.62} \\
		
		\bottomrule
	\end{tabular}
\end{table*}

Table~\ref{tab:heat_stress_results} compares heat-stress recognition methods on the HSCow dataset. SlowFast and Video Swin achieve Top-1 accuracies of 70.18\% and 59.65\%, respectively. The relatively low performance of these high-capacity architectures suggests an increased risk of overfitting with the limited training data available in this setting. TSN and TSM achieve 78.07\% and 76.32\%, respectively, indicating that sparse temporal modeling alone is insufficient to resolve the subtle, high-frequency patterns associated with heat stress. ActionDiff uses diffusion features and obtains a Top-1 accuracy of 76.32\%. However, it does not include a task-specific mechanism that aligns general diffusion features with localized motion cues such as flank undulations. Using $T=8$ and $\Delta t=8$, TSDF-Net achieves the best performance, with a Top-1 accuracy of 87.72\% and a macro-F1 score of 87.62\%. These results exceed those of Video-FocalNet by 5.26 pp and 5.21 pp, respectively, supporting the complementary roles of TSAM in suppressing temporal noise and DSP in incorporating motion priors under limited training data.

\subsection{Ablation Studies}

\begin{table}
	\centering
	\caption{Ablation of VDA-Net components on cross-camera ReID (\%). Changes are absolute percentage-point differences from the preceding row.}
	\label{tab:reid_ablation}
	\small
	\setlength{\tabcolsep}{4pt}
	\begin{tabular}{lcccc}
		\toprule
		\textbf{Configuration} & \textbf{Rank-1 $\uparrow$} & \textbf{Rank-5 $\uparrow$} & \textbf{Rank-10 $\uparrow$} & \textbf{mAP $\uparrow$} \\
		\midrule
		Baseline & 96.03 & 98.72 & 99.05 & 96.69 \\
		$+$ Outlook & 97.50 & 99.29 & 99.33 & 97.54 \\
		\multicolumn{1}{r}{\scriptsize\textit{vs. previous}} & \gain{1.47\,\mathrm{pp}} & \gain{0.57\,\mathrm{pp}} & \gain{0.28\,\mathrm{pp}} & \gain{0.85\,\mathrm{pp}} \\
		\addlinespace[2pt]
		$+$ VDT & \textbf{99.83} & \textbf{99.98} & \textbf{99.98} & \textbf{99.67} \\
		\multicolumn{1}{r}{\scriptsize\textit{vs. previous}} & \gain{2.33\,\mathrm{pp}} & \gain{0.69\,\mathrm{pp}} & \gain{0.65\,\mathrm{pp}} & \gain{2.13\,\mathrm{pp}} \\
		\bottomrule
	\end{tabular}
\end{table}

Table~\ref{tab:reid_ablation} details the impact of VDA-Net's internal modules on the ReID benchmark. Introducing the Outlook block yields improvements of 1.47 pp in Rank-1 and 0.85 pp in mAP, validating that multi-scale localized aggregation effectively captures visible textures under partial occlusions. The subsequent integration of the Viewpoint Decoupling Transformer (VDT) brings further gains of 2.33 pp in Rank-1 and 2.13 pp in mAP, confirming our hypothesis that explicitly disentangling viewpoint-dependent variations is critical for cross-camera identity matching.

\begin{table}
	\centering
	\caption{Ablation of CowTrack components on global tracking (\%). Changes are absolute percentage-point differences from the preceding row.}
	\label{tab:tracking_ablation}
	\small
	\setlength{\tabcolsep}{6pt}
	\begin{tabular}{lccc}
		\toprule
		\textbf{Configuration} & \textbf{HOTA $\uparrow$} & \textbf{MOTA $\uparrow$} & \textbf{IDF1 $\uparrow$} \\
		\midrule
		Baseline & 77.49 & 82.76 & 81.40 \\
		$+$ VDA-Net (Static) & 81.67 & \textbf{89.50} & 85.55 \\
		\multicolumn{1}{r}{\scriptsize\textit{vs. previous}} & \gain{4.18\,\mathrm{pp}} & \gain{6.74\,\mathrm{pp}} & \gain{4.15\,\mathrm{pp}} \\
		\addlinespace[2pt]
		$+$ Online Adaptation & \textbf{83.48} & \textbf{89.50} & \textbf{87.72} \\
		\multicolumn{1}{r}{\scriptsize\textit{vs. previous}} & \gain{1.81\,\mathrm{pp}} & \nochange{0.00\,\mathrm{pp}} & \gain{2.17\,\mathrm{pp}} \\
		\bottomrule
	\end{tabular}
\end{table}

For the tracking pipeline (Table~\ref{tab:tracking_ablation}), replacing the static baseline extractor with VDA-Net increases HOTA by 4.18 pp and MOTA by 6.74 pp, translating view-invariant embeddings into reliable data association. Causal online adaptation further increases IDF1 by 2.17 pp while maintaining MOTA at 89.50\%. This improvement in IDF1 suggests that dynamic self-supervision enhances long-term identity consistency against temporal drift.

\begin{table}
	\centering
	\caption{Ablation studies of TSDF-Net components (\%). Changes are absolute percentage-point differences from the preceding row.}
	\label{tab:tsdf_ablation}
	
	\small
	\setlength{\tabcolsep}{4pt}
	
	\begin{tabular}{lcccc}
		\toprule
		\textbf{Configuration} & \textbf{Top-1 $\uparrow$} & \textbf{Precision $\uparrow$} & \textbf{Recall $\uparrow$} & \textbf{F1 $\uparrow$} \\
		\midrule
		
		Baseline 
		& 82.46 & 83.06 & 82.57 & 82.41 \\
		
		$+$ TSAM 
		& 83.30 & 87.33 & 83.62 & 82.95 \\
		
		\multicolumn{1}{r}{\scriptsize\textit{vs. previous}} 
		& \gain{0.84\,\mathrm{pp}} 
		& \gain{4.27\,\mathrm{pp}} 
		& \gain{1.05\,\mathrm{pp}} 
		& \gain{0.54\,\mathrm{pp}} \\
		
		\addlinespace[2pt]
		
		$+$ DSP 
		& \textbf{87.72} 
		& \textbf{89.35} 
		& \textbf{87.90} 
		& \textbf{87.62} \\
		
		\multicolumn{1}{r}{\scriptsize\textit{vs. previous}} 
		& \gain{4.42\,\mathrm{pp}} 
		& \gain{2.02\,\mathrm{pp}} 
		& \gain{4.28\,\mathrm{pp}} 
		& \gain{4.67\,\mathrm{pp}} \\
		
		\bottomrule
	\end{tabular}
	
\end{table}

For the recognition pipeline (Table~\ref{tab:tsdf_ablation}), adding TSAM improves the baseline's Top-1 accuracy by 0.84 pp and precision by 4.27 pp, indicating that temporal smoothing successfully mitigates the interference of noisy frames. The integration of DSP further elevates Top-1 accuracy to 87.72\%. The combined use of both modules yields an overall Top-1 improvement of 5.26 pp over the baseline, suggesting that TSAM provides a stabilized feature sequence that enables the gated fusion mechanism in DSP to incorporate diffusion semantics more accurately.

\subsection{Multi-View Studies}
The comparison between single-camera and multi-camera setups (Table~\ref{tab:camera_setup_comparison}) indicates the advantage of our multi-view representation. The multi-camera setup outperformed the single-camera setup across all evaluation metrics, with significant differences in the magnitude of improvement across different metrics. For Rank-k metrics, Rank-1 improved from 98.30\% to 99.83\%, an increase of 1.53 percentage points, indicating that multi-camera perspective information helps improve the accuracy of first-place matching. Meanwhile, mAP improved from 88.93\% to 99.67\%, an increase of 10.74 percentage points, significantly higher than the improvement seen in Rank-k metrics. These results demonstrate that the multi-camera setup not only improves the model's ability to hit the top few positions but also enhances the global ranking consistency of samples with the same identity, as measured by mAP. Compared to the single-camera setup, multi-camera perspective information enhances the model's adaptability to cross-viewpoint changes and occluded samples, making the distribution of samples with the same identity more compact in the feature space, thereby improving overall retrieval quality.

\begin{table}
	\centering
	\caption{Comparison of cross-camera ReID performance between single-camera and multi-camera setups (\%).}
	\label{tab:camera_setup_comparison}
	\begin{tabular}{lcccc}
		\toprule
		\textbf{Setup} & \textbf{Rank-1 $\uparrow$} & \textbf{Rank-5 $\uparrow$} & \textbf{Rank-10 $\uparrow$} & \textbf{mAP $\uparrow$} \\
		\midrule
		Single-CAM & 98.30 & 99.20 & 99.73 & 88.93 \\
		Multi-CAM & \textbf{99.83} & \textbf{99.98} & \textbf{99.98} & \textbf{99.67} \\
		\midrule
		Gain (pp) & \gain{1.53} & \gain{0.78} & \gain{0.25} & \gain{10.74} \\
		\bottomrule
	\end{tabular}
\end{table}

\subsection{Time Sampling Sensitivity}

\begin{figure*}
	\centering
	\includegraphics[width=0.8\linewidth]{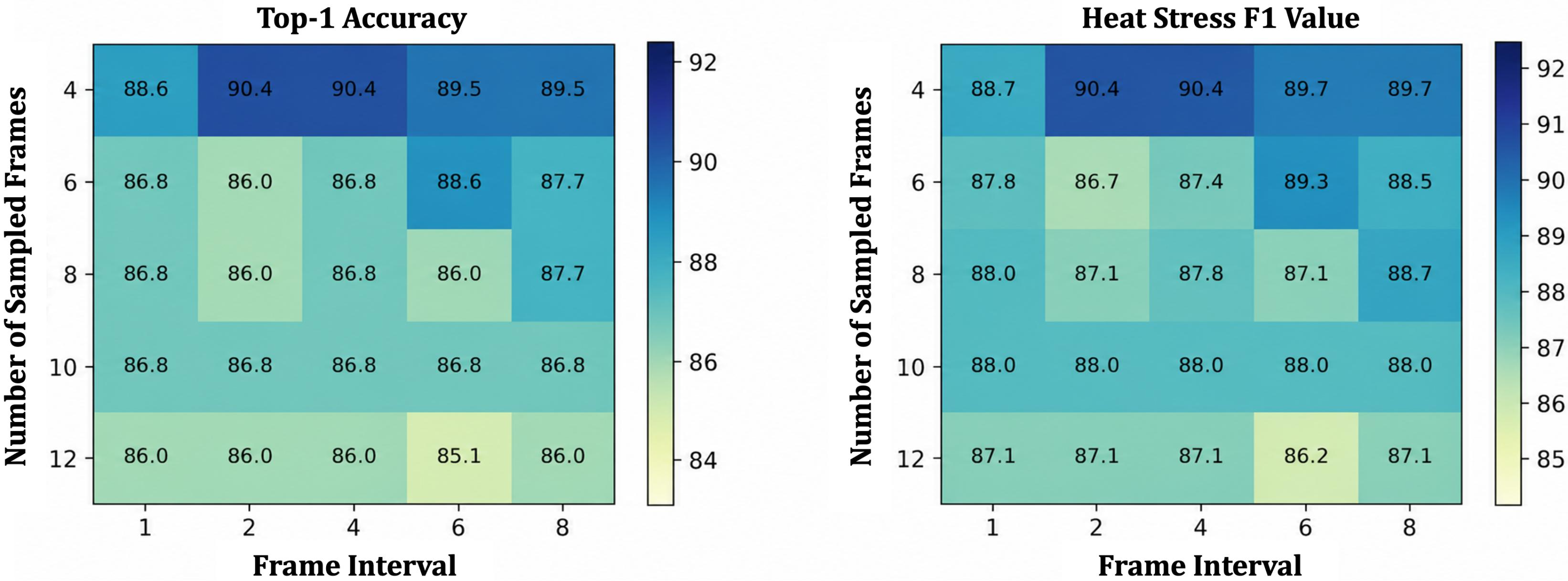}
	\caption{Heatmap visualization of TSDF-Net performance sensitivity to temporal sampling parameters (input frame count $T$ and frame interval $\Delta t$) in terms of Top-1 accuracy (left) and Heat Stress F1 score (right).}
	\label{fig:time_sensitive_heatmap}
\end{figure*}

To evaluate TSDF-Net's temporal robustness, we vary the number of input frames $T \in \{4, 6, 8, 10, 12\}$ and the frame interval $\Delta t \in \{1, 2, 4, 6, 8\}$ (Fig.~\ref{fig:time_sensitive_heatmap}, Table~\ref{tab:temporal_sampling_sensitivity}). Fig.~\ref{fig:time_sensitive_heatmap} reports the complete $5 \times 5$ grid, whereas Table~\ref{tab:temporal_sampling_sensitivity} lists representative configurations with additional class-level metrics. Across all 25 configurations, the mean Top-1 accuracy is 87.23\% with a standard deviation of 1.43 pp, and the mean Heat Stress F1 score is 88.07\% with a standard deviation of 1.10 pp.

For $T=4$, the highest Top-1 values in the complete grid occur at $\Delta t=2$ and $\Delta t=4$, both approximately 90.4\% after rounding. However, we assess temporal sampling configurations using the complete metric profile rather than Top-1 alone; Table~\ref{tab:temporal_sampling_sensitivity} therefore presents representative settings for comparing overall and heat-stress-specific performance. The effect of $\Delta t$ depends on $T$: the results are identical across intervals for $T=10$, whereas modest variations are observed for the other clip lengths. Heat Stress Recall reaches 98.21\% for $T=8$ with $\Delta t=1$ or 8 and for $T=10$ across all intervals. Because recall is particularly important for minimizing missed heat-stress cases, we select $T=8$ and $\Delta t=8$ as the final configuration. This setting yields a Top-1 accuracy of 87.72\%, a macro-F1 score of 87.62\%, and a Heat Stress Recall of 98.21\%, and is used for all reported TSDF-Net results.

\begin{table*}
	\centering
	\caption{Sensitivity analysis of selected representative temporal sampling configurations, parameterized by the input frame count $T$ and frame interval $\Delta t$, on TSDF-Net performance (\%). The complete results for all 25 configurations are shown in Fig.~\ref{fig:time_sensitive_heatmap}. Precision, Recall, and F1 are macro-averaged across both classes. Heat Stress F1 and Heat Stress Recall treat the heat stress category as the positive class. For $T=10$, all interval results are identical, hence only one representative row is listed.}
	\label{tab:temporal_sampling_sensitivity}
	\begin{tabular}{@{}cccccccc@{}}
		\toprule
		\textbf{Input Frames $T$} & \textbf{Frame Interval $\Delta t$} & \textbf{Top-1 $\uparrow$} & \textbf{Precision $\uparrow$} & \textbf{Recall $\uparrow$} & \textbf{F1 $\uparrow$} & \textbf{\begin{tabular}[c]{@{}c@{}}Heat Stress F1 $\uparrow$\end{tabular}} & \textbf{\begin{tabular}[c]{@{}c@{}}Heat Stress Recall $\uparrow$\end{tabular}} \\
		\midrule
		4 & 1 & 88.60 & 88.68 & 88.64 & 88.60 & 88.70 & 91.07 \\
		4 & 8 & 89.47 & 89.63 & 89.53 & 89.47 & 89.66 & 92.86 \\
		\addlinespace
		6 & 1 & 86.84 & 88.17 & 87.01 & 86.76 & 87.80 & 96.43 \\
		6 & 8 & 87.72 & 88.83 & 87.87 & 87.66 & 88.52 & 96.43 \\
		\addlinespace
		8 & 1 & 86.84 & 88.74 & 87.04 & 86.72 & 88.00 & 98.21 \\
		8 & 8 & 87.72 & 89.35 & 87.90 & 87.62 & 88.71 & 98.21 \\
		\addlinespace
		10 & --- & 86.84 & 88.74 & 87.04 & 86.72 & 88.00 & 98.21 \\
		\addlinespace
		12 & 1 & 85.96 & 87.53 & 86.14 & 85.86 & 87.10 & 96.43 \\
		12 & 8 & 85.96 & 87.53 & 86.14 & 85.86 & 87.10 & 96.43 \\
		\bottomrule
	\end{tabular}
\end{table*}

\subsection{Qualitative Analysis and Interpretability}

\subsubsection{Robust Tracking under Severe Occlusion}
\begin{figure*}
	\centering
	\includegraphics[width=0.95\linewidth]{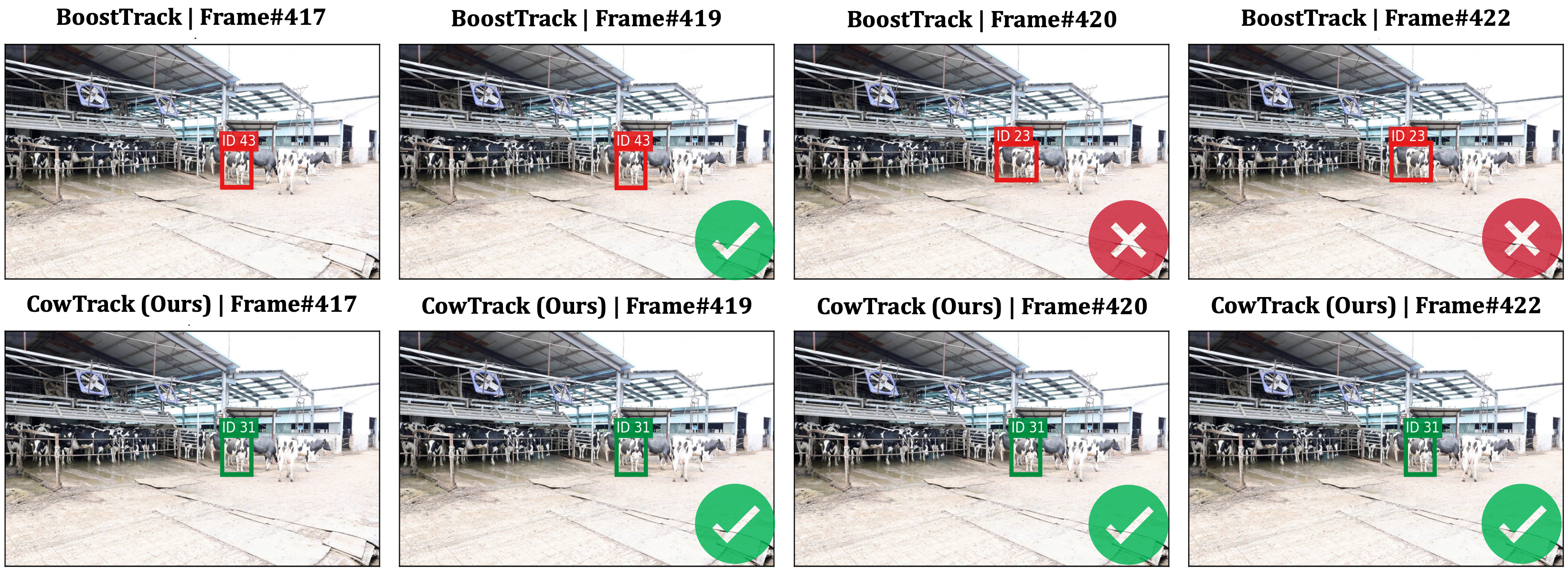}
	\caption{Qualitative comparison of tracking performance under severe occlusion. The baseline method (top) suffers from an ID switch (from ID 43 to ID 23), whereas our CowTrack (bottom) successfully maintains the correct identity (ID 31).}
	\label{fig:occlusion_tracking}
\end{figure*}

A qualitative comparison under severe occlusion is shown in Fig.~\ref{fig:occlusion_tracking}. During continuous tracking from frames 417 to 422, the target cow was obstructed by surrounding cattle and fences, resulting in a significant reduction in its visible area. BoostTrack labeled the target as ID 43 in frames 417 and 419, but reassigned it to ID 23 after frame 420, exhibiting a clear identity switch. In contrast, CowTrack consistently maintained the target as ID 31 in the same video segment, without any identity switching. These results demonstrate that the discriminative appearance features provided by VDA-Net can effectively improve the accuracy of identity association during tracking, reduce ID switching and trajectory breakage under occlusion conditions, and provide more stable trajectory input for subsequent individual-level heat stress behavior recognition.

\subsubsection{Feature Space Analysis via t-SNE}

\begin{figure*}
	\centering
	\includegraphics[width=0.8\linewidth]{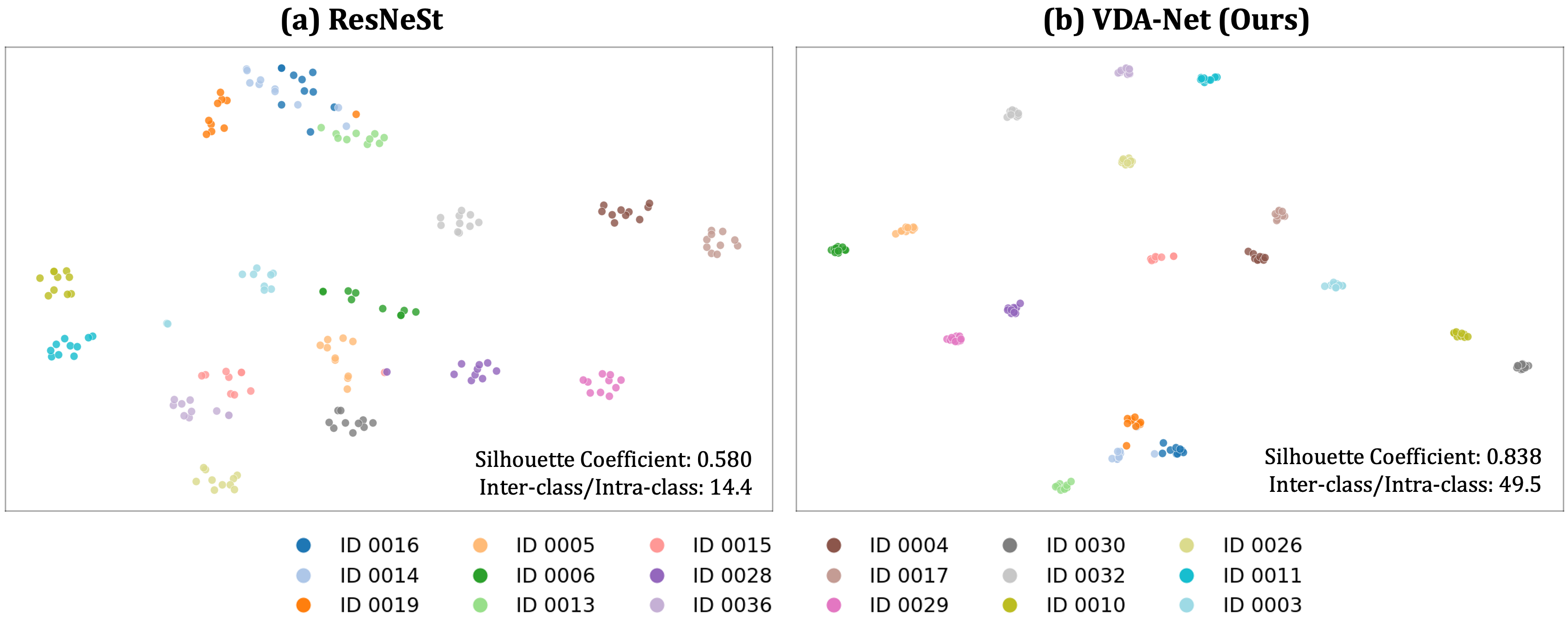}
	\caption{t-SNE feature space visualization of hard ReID samples in the MVCow dataset. Compared to the baseline (left), our VDA-Net (right) yields tighter intra-class clustering and larger inter-class margins.}
	\label{fig:reid_tsne}
\end{figure*}

To evaluate the representational capacity of our framework, we first visualize the feature space of the identity re-identification module using t-SNE. As illustrated in Fig.~\ref{fig:reid_tsne}, VDA-Net yields tighter intra-class clustering and larger inter-class margins compared to the baseline. Specifically, the baseline ResNeSt model exhibits significant overlap among challenging identities, resulting in blurred decision boundaries that commonly induce retrieval and ranking errors. In contrast, VDA-Net produces highly concentrated clusters for samples of the same identity alongside substantially widened inter-class intervals. Quantitatively, the silhouette coefficient improves from 0.580 to 0.838, and the inter-to-intra-class distance ratio increases from 14.4 to 49.5. This validates that the Outlook attention module successfully accentuates fine-grained local patterns, while the VDT viewpoint decoupling module effectively mitigates the interference of viewpoint variations, ultimately yielding highly discriminative and perspective-robust identity representations.

\begin{figure*}
	\centering
	\includegraphics[width=0.8\linewidth]{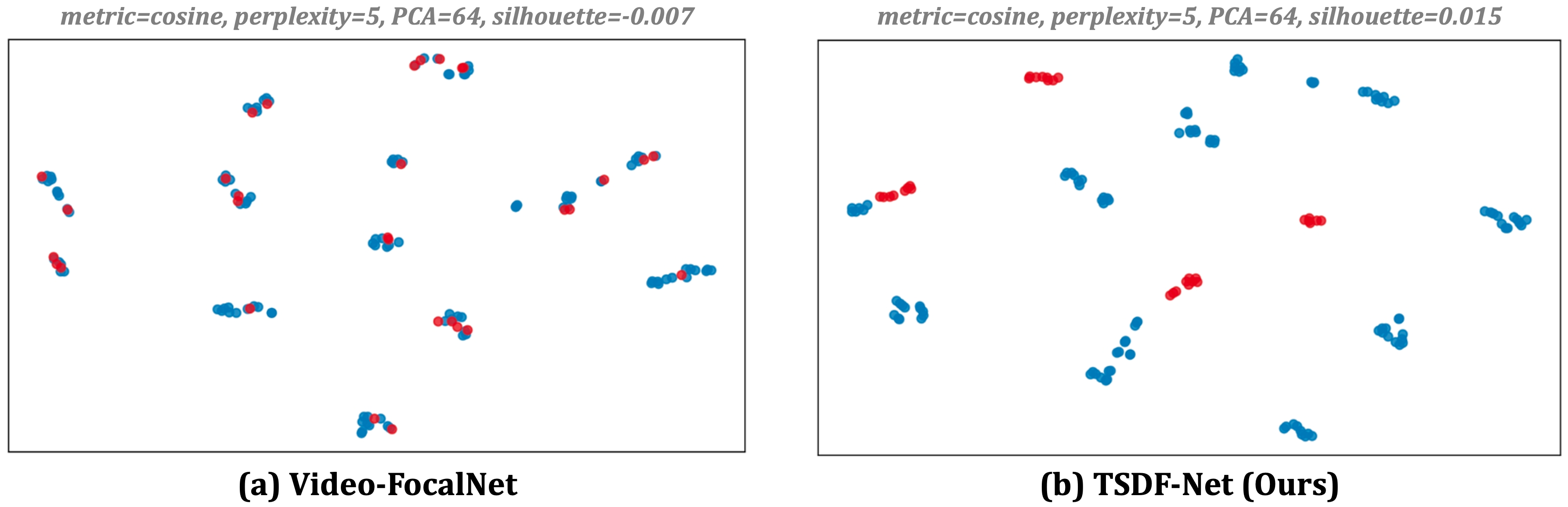}
	\caption{Comparison of feature space distributions using t-SNE. (a) The Video-FocalNet baseline; (b) Our TSDF-Net, which demonstrates clearer decision boundaries between normal and heat-stressed samples.}
	\label{fig:tsdf_tsne}
\end{figure*}

Beyond identity modeling, we further analyze the feature discriminability of the downstream behavior recognition network. The fusion features of all test samples, obtained via the \(h_{\mathrm{fuse}}\) operation, are projected onto a two-dimensional plane. Fig.~\ref{fig:tsdf_tsne} compares the feature distributions generated by the Video-FocalNet baseline and our TSDF-Net. The baseline feature space demonstrates severe overlap between normal and heat-stressed samples, lacking a discernible classification boundary. Conversely, TSDF-Net successfully separates these two behavioral states into well-defined, concentrated clusters. The significantly expanded inter-class distance indicates an enhancement in feature discriminability, demonstrating that TSDF-Net reliably disentangles complex dynamic features for accurate heat stress diagnosis.

\subsubsection{Cross-Camera Identity Retrieval}
\begin{figure*}
	\centering
	\includegraphics[width=0.95\linewidth]{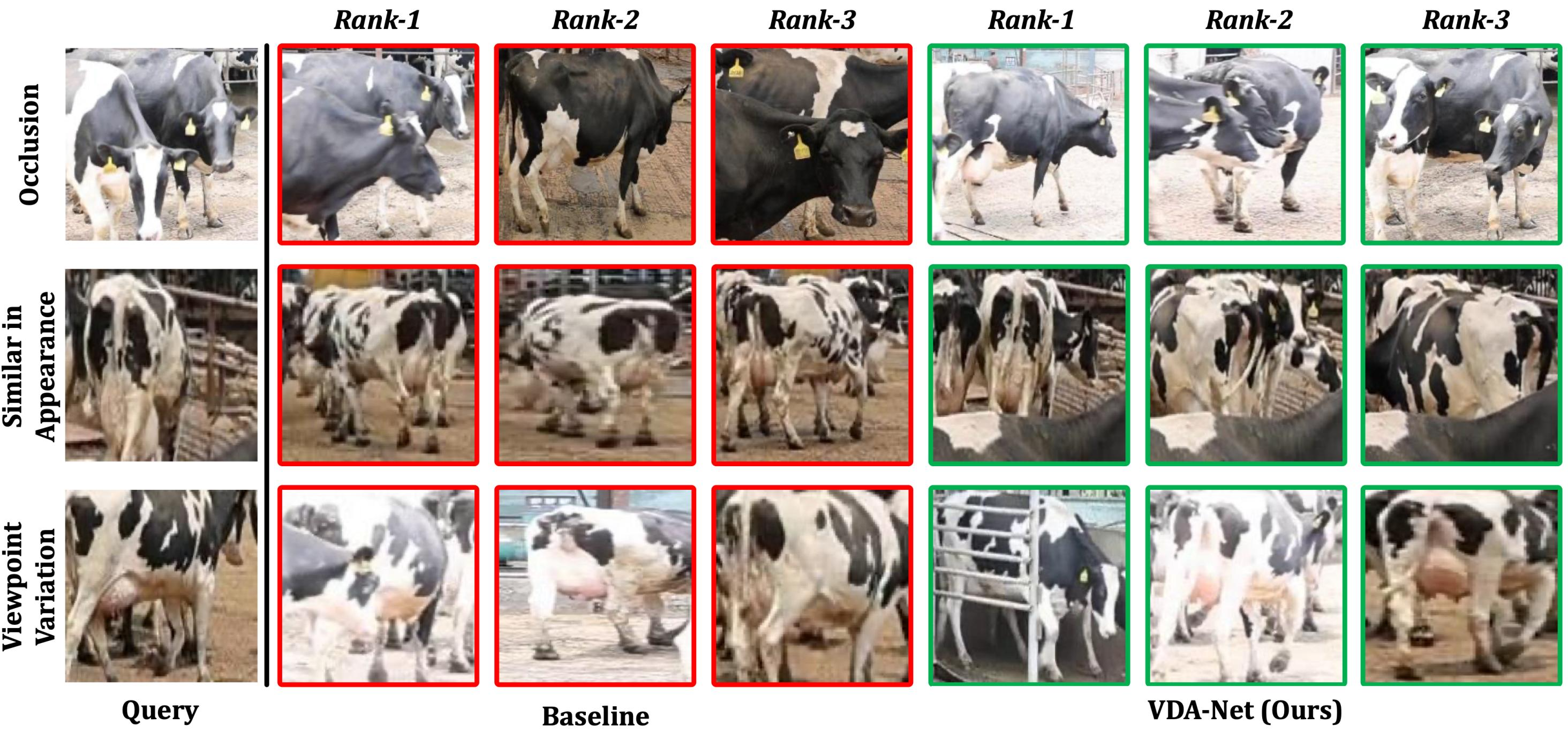}
	\caption{Cross-camera retrieval rankings produced by the baseline and VDA-Net in challenging cases involving occlusion and viewpoint changes.}
	\label{fig:reid_visualization}
\end{figure*}

Qualitative ReID examples are shown in Fig.~\ref{fig:reid_visualization}. The query samples cover scenarios involving occlusion, viewpoint changes, and highly similar patterns. Each query image is accompanied by the top-three retrieval results returned by the model. As shown in Fig.~\ref{fig:reid_visualization}, the BoostTrack baseline produces incorrect matches under occlusion, viewpoint changes, and highly similar individual appearances. In contrast, VDA-Net retrieves the correct identity at Rank-1 in all three examples, indicating stronger fine-grained feature discrimination and greater robustness to viewpoint changes.

\subsubsection{Spatiotemporal Attention and Interpretability}

\begin{figure*}
	\centering
	\includegraphics[width=0.95\linewidth]{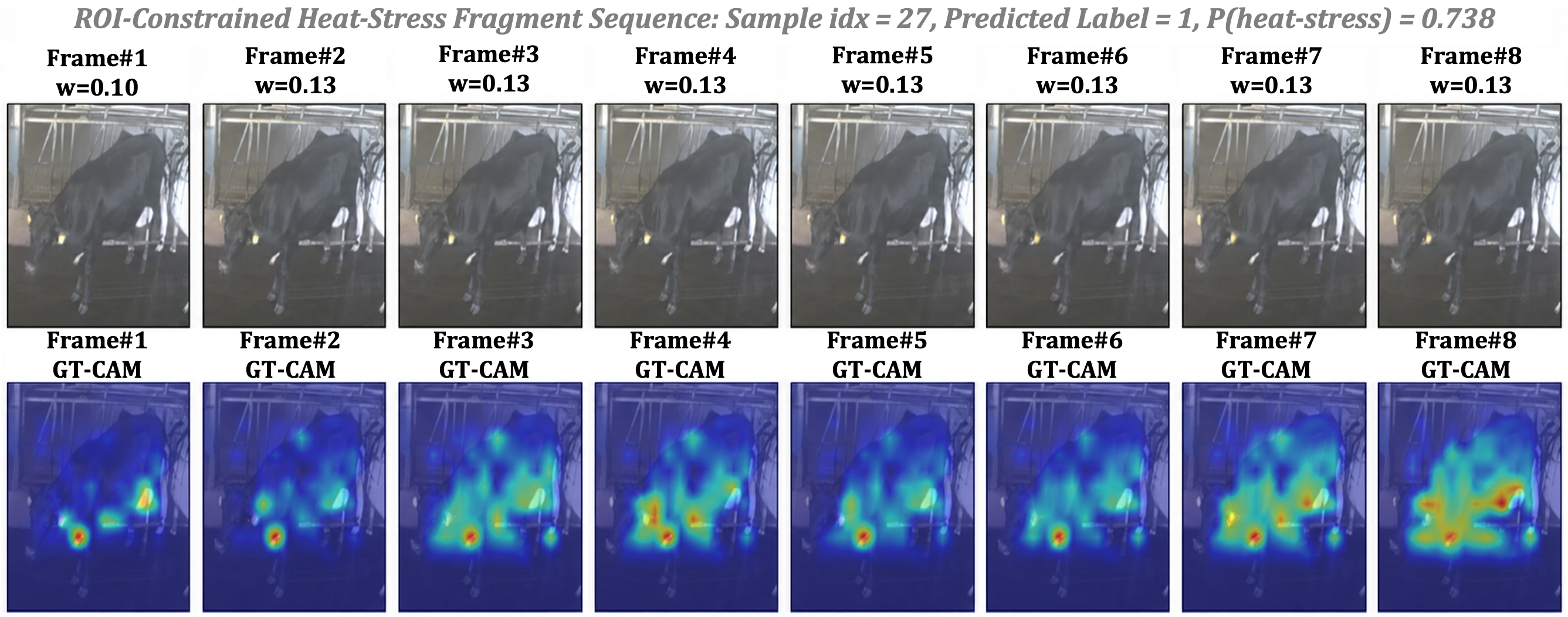}
	\caption{Grad-CAM visualization of a heat stress video sequence constrained by the cow body ROI. The heatmaps demonstrate that our network correctly focuses on physiological indicators such as the flank undulations.}
	\label{fig:gradcam}
\end{figure*}

To interpret the decision-making mechanism of our behavior recognition network, we visualize the spatiotemporal activations using Grad-CAM. Fig.~\ref{fig:gradcam} displays eight consecutively sampled frames from a heat stress video sequence (top) and their corresponding ROI-constrained heatmaps (bottom). Temporally, the attention weight distribution remains well-balanced across the sequence, fluctuating narrowly between 0.10 and 0.13, which closely aligns with the uniform average of 0.125. This consistency indicates that the model avoids single-frame bias and comprehensively exploits sequence-level temporal context for classification. Spatially, the heatmaps consistently highlight critical physiological areas, including the abdominal contour, lower torso, forelimbs, and head. The absence of spatial drift over time demonstrates the capacity of TSDF-Net to stably capture dynamic, fine-grained features intrinsically linked to heat stress.

\begin{figure}
	\centering
	\includegraphics[width=0.95\linewidth]{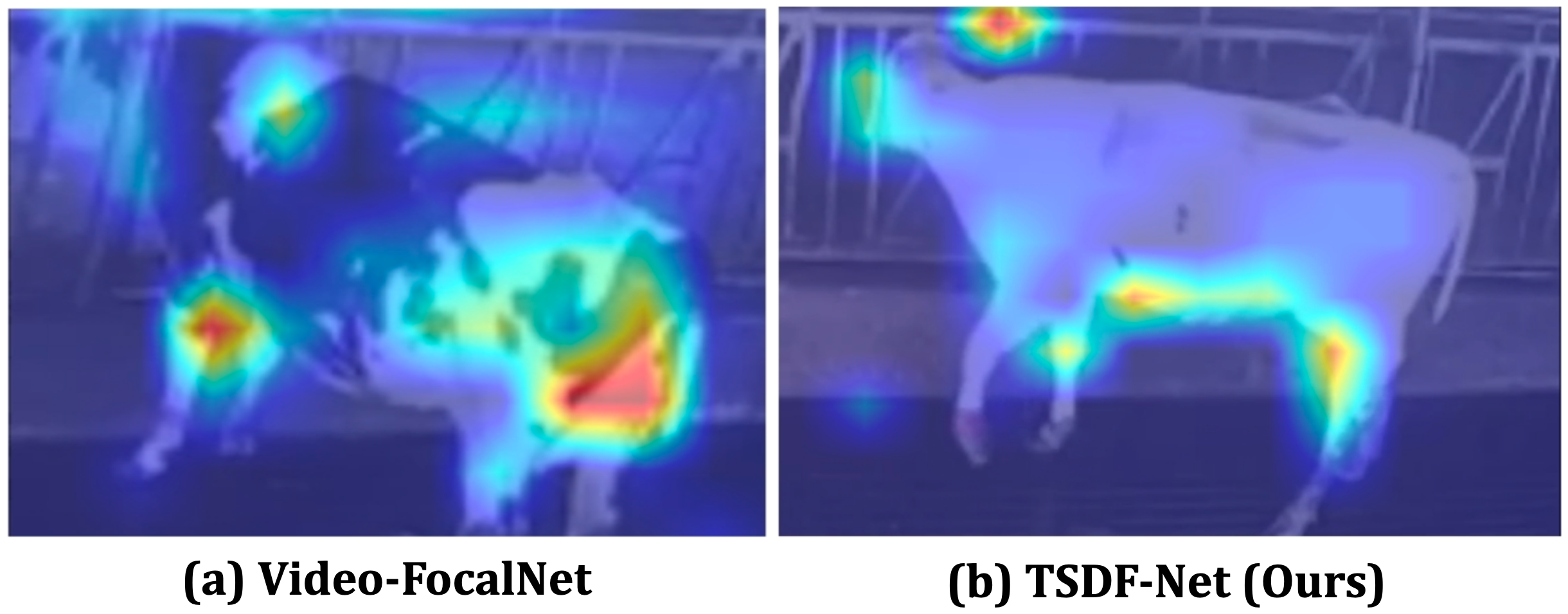}
	\caption{Attention heatmap comparison. Compared to the baseline (left), TSDF-Net (right) effectively concentrates on critical physiological areas (e.g., flank undulations) while avoiding background distractions.}
	\label{fig:tsdf_attention}
\end{figure}

Building upon this spatiotemporal stability, we further compare the specific attention distribution of TSDF-Net against the Video-FocalNet baseline. As illustrated in Fig.~\ref{fig:tsdf_attention}, the baseline model primarily attends to less informative regions such as the rump and neck, failing to capture the key physiological indicators of heat stress. In sharp contrast, TSDF-Net exhibits precise and concentrated activations on the mouth, nose, abdomen, and legs. These highly localized regions directly correspond to the hallmark behaviors of heat stress, namely head shaking, abdominal undulation, and postural instability. Furthermore, TSDF-Net maintains a persistent focus on the foreground subject without dispersing attention into the environment, validating that the TSAM module effectively suppresses background interference and camera jitter at the feature level to ensure reliable interpretability.

\subsubsection{Temporal Noise Suppression}

\begin{figure*}
	\centering
	\includegraphics[width=0.8\linewidth]{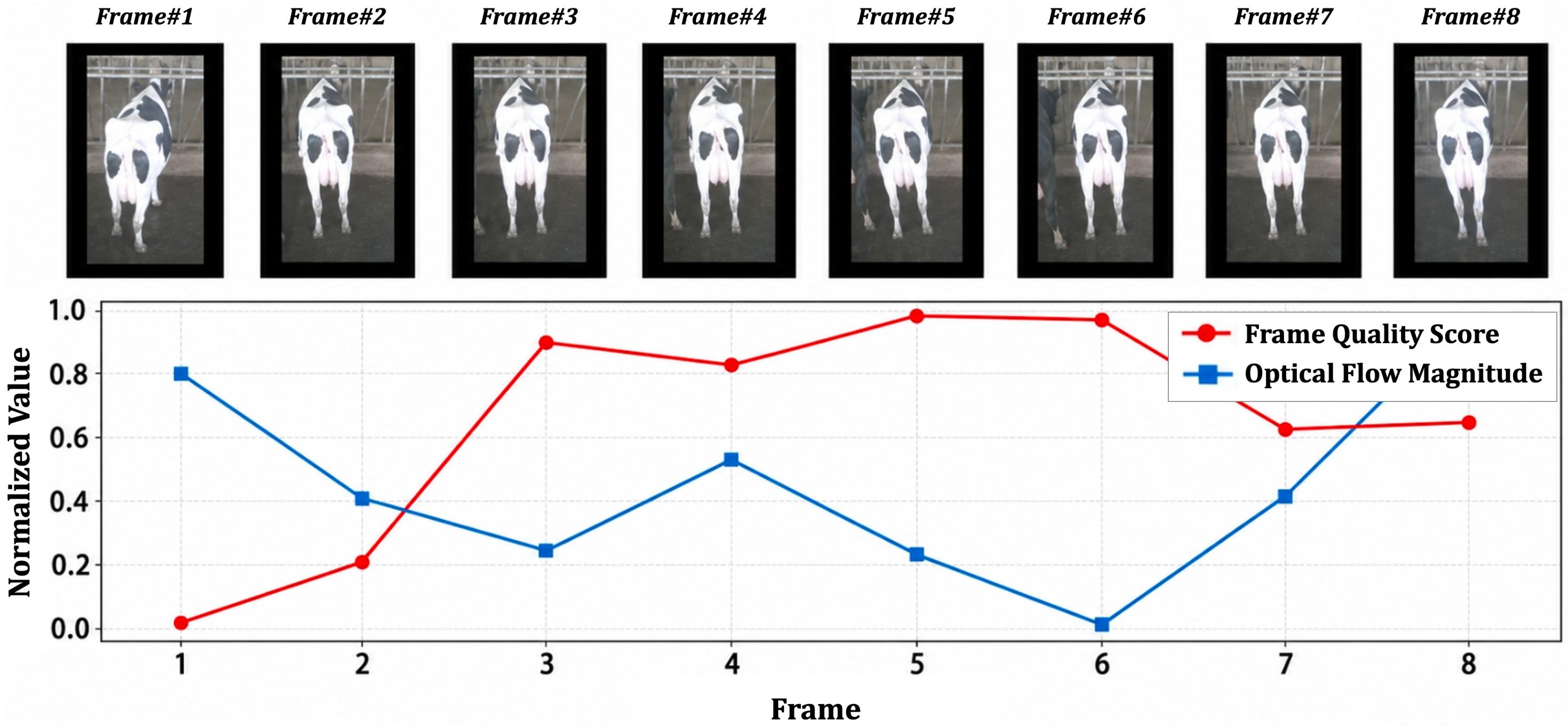}
	\caption{Temporal sequence comparison between frame quality scores generated by TSAM (red) and optical flow magnitude (blue), illustrating the model's ability to adaptively down-weight high-jitter frames.}
	\label{fig:tempo_comparison}
\end{figure*}

A video segment containing significant camera jitter was selected from the test set. The frame quality scores (\(q_t\)) output by the TSAM frame quality assessment head were plotted as curves along the time axis and compared with the inter-frame optical flow variation amplitude. The results are shown in Fig.~\ref{fig:tempo_comparison}. It can be seen that at the locations where the inter-frame optical flow amplitude shows a significant peak, the frame quality score output by TSAM drops significantly, indicating that the frame quality assessment head can effectively identify jittery frames and reduce their weight, thereby reducing interference from noisy frames.

\subsection{Integrated Pipeline Evaluation}

\begin{figure*}
	\centering
	\includegraphics[width=0.8\linewidth]{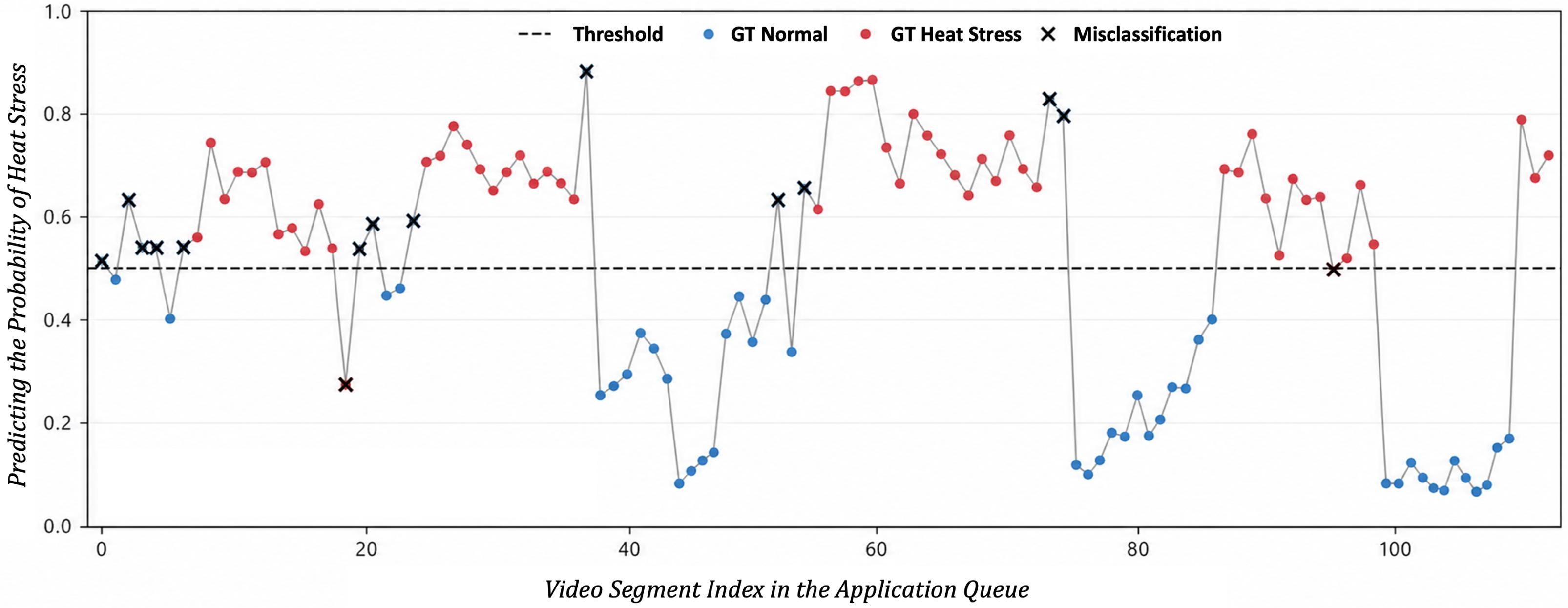}
	\caption{Continuous heat stress prediction probability curve generated by the CowCare framework in a simulated real-world deployment queue.}
	\label{fig:continuous_scan}
\end{figure*}

To demonstrate the practical viability of the unified CowCare framework, we conducted a continuous queue scanning test, simulating a real-world deployment pipeline where CowTrack sequentially feeds uncurated tracklets to TSDF-Net. As shown in the prediction probability curve (Fig.~\ref{fig:continuous_scan}), the cascaded framework maintains robust performance, correctly identifying heat-stressed cattle with a 98.21\% recall rate. This suggests that the upstream viewpoint-robust tracking and downstream anti-jitter classification function synergistically, successfully bridging the gap between raw video pixels and reliable livestock health diagnostics.

\section{Conclusion}
\label{sec:conclusion}
In this study, we address the critical yet underexplored challenge of non-invasive heat stress monitoring in complex commercial dairy farms. We propose CowCare, a unified visual framework that connects multi-object tracking and video-based behavior analysis into a cohesive diagnostic pipeline. By introducing the VDA-Net appearance model and causal online self-supervision, the upstream CowTrack module achieves robust identity preservation, effectively mitigating the physical challenges of severe occlusion and cross-camera viewpoint distortion. Furthermore, by feeding these stable tracklets into the downstream TSDF-Net, our framework mitigates the effects of high-frequency physical vibrations and limited training data through temporal stabilization and diffusion semantic priors. Extensive evaluations on our newly constructed MVCow and HSCow datasets demonstrate that CowCare achieves state-of-the-art performance in both identity preservation and fine-grained behavior recognition. Future work will focus on optimizing the computational efficiency of the framework for edge-device deployment and extending its application to broader veterinary monitoring tasks, such as estrus and lameness detection across diverse agricultural environments.

\section*{Acknowledgment}
This work was supported by the Beijing Smart Agriculture Innovation Consortium Project (BAIC10-2026).

 \printcredits

\bibliographystyle{cas-model2-names}

\bibliography{cas-refs}

\end{document}